\documentclass[journal]{IEEEtran}
\usepackage{cite}
\usepackage{amsmath,amssymb,amsfonts}
\usepackage{algorithmic}
\usepackage{graphicx}
\usepackage{subcaption}
\usepackage{textcomp}
\usepackage{xcolor}
\usepackage{hyperref}
\usepackage{booktabs}
\usepackage{threeparttable}
\usepackage{multirow}
\usepackage{array}
\usepackage{float}
\usepackage{flushend}
\usepackage{amsthm}
\usepackage{enumitem}
\usepackage{pifont}

\newcommand{\moxy}[1]{\textcolor{black}{#1}}
\newcommand{\moa}[1]{\textcolor{black}{#1}}
\newcommand{\gjt}[1]{\textcolor{black}{#1}}
\newcommand{\wzf}[1]{\textcolor{black}{#1}}
\newcommand{\kal}[1]{\textcolor{black}{#1}}

\begin{document}
\title{SparScene: Efficient Traffic Scene Representation via Sparse Graph Learning for Large-Scale Trajectory Generation}
\author{Xiaoyu~Mo,~\IEEEmembership{Member,~IEEE,}
        and~Jintian~Ge,
        and~Zifan~Wang,~\IEEEmembership{Member,~IEEE,}
        and~Chen~Lv,~\IEEEmembership{Senior Member,~IEEE,}
        and~Karl~H.~Johansson,~\IEEEmembership{Fellow,~IEEE}
\thanks{Xiaoyu Mo, Zifan Wang, and Karl H. Johansson are with the Division of Decision and Control Systems, School of Electrical Engineering and Computer Science, KTH Royal Institute of Technology, Stockholm, Sweden. (e-mail: xiaoyu006@e.ntu.edu.sg, zifanw@kth.se, kallej@kth.se)}
\thanks{Jintian Ge is currently with iFLYTEK Co., Ltd., Shanghai, China. (e-mail: jintian001@e.ntu.edu.sg).}
\thanks{Chen Lv is with the School of Mechanical and Aerospace Engineering, Nanyang Technological University, 639798, Singapore. (e-mail: lyuchen@ntu.edu.sg).}
}
\maketitle

\begin{abstract}  
Multi-agent trajectory generation is a core problem for autonomous driving and intelligent transportation systems. 
However, efficiently modeling the dynamic interactions between numerous road users and infrastructures in complex scenes remains an open problem. 
Existing methods typically employ \moa{distance-based or fully connected dense graph structures} to capture interaction information, which not only introduces a large number of redundant edges but also requires complex and heavily parameterized networks for encoding, thereby resulting in low training and inference efficiency, limiting scalability to large and complex traffic scenes.
To overcome the limitations of existing methods, we propose SparScene, a sparse graph learning framework designed for efficient and scalable traffic scene representation. 
Instead of relying on distance thresholds, SparScene leverages the lane graph topology to construct structure-aware sparse connections between agents and lanes, enabling efficient yet informative scene graph representation.
SparScene adopts a lightweight graph encoder that efficiently aggregates agent–map and agent–agent interactions, yielding compact scene representations with \wzf{substantially improved} efficiency and scalability. 
On the motion prediction benchmark of the Waymo Open Motion Dataset (WOMD), SparScene achieves competitive performance \wzf{with} remarkable efficiency. It generates trajectories for more than 200 agents in a scene within 5 ms and scales to more than 5,000 agents and 17,000 lanes with merely 54 ms of inference time with a GPU memory of 2.9 GB, highlighting its superior scalability for large-scale traffic scenes.
\end{abstract}


\IEEEpeerreviewmaketitle

\begin{figure}[t]
    \centering
    \includegraphics[trim={0cm 0cm 0cm 0cm}, clip, width=0.48\textwidth]{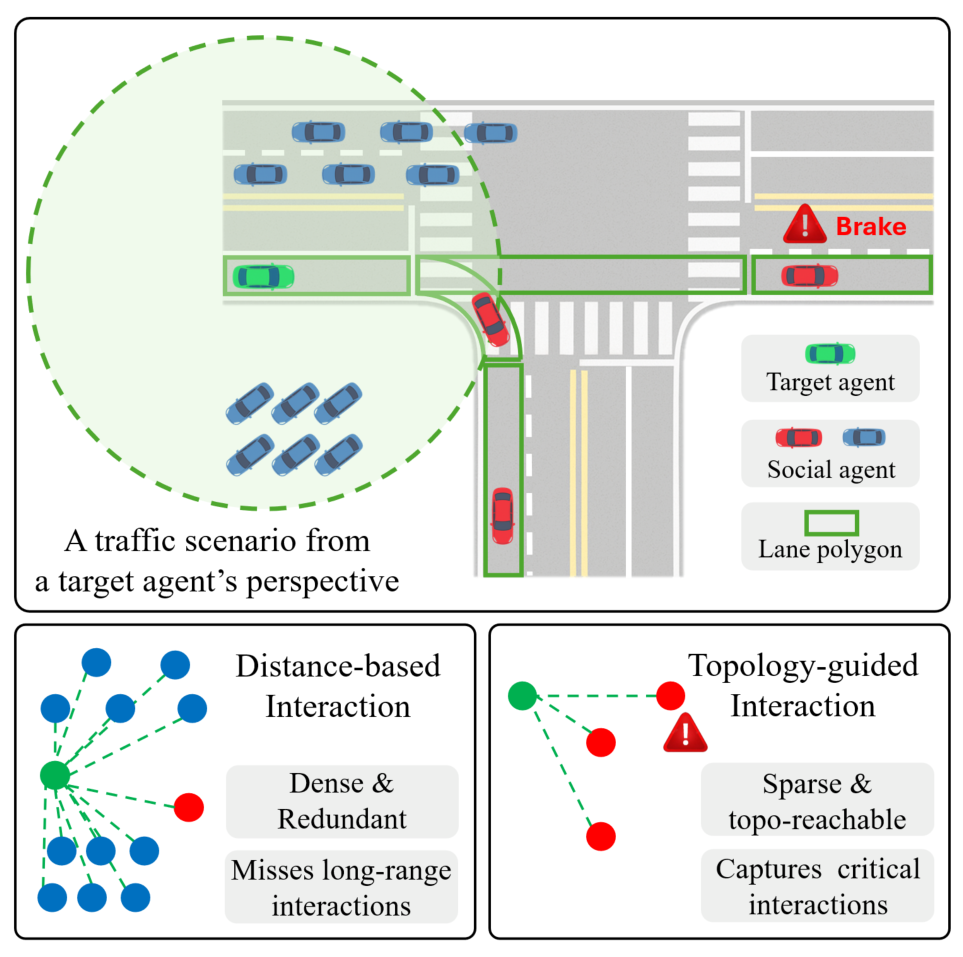}
    \vspace{-2mm}
    \caption{SparScene proposes a topology-guided interaction modeling method that selectively models topology-consistent interactions with strong behavioral relevance, resulting in sparse and interpretable interaction graphs.
    \textit{Upper:} A local traffic scene from the perspective of the target agent.
    \textit{Bottom left:} Distance-based interaction modeling produces dense and redundant connections and may miss behaviorally important long-range interactions.
    \textit{Bottom right:} Topology-guided interaction modeling in SparScene yields sparse, interpretable interaction graphs by capturing topology-consistent interactions beyond distance constraints.
    Note: SparScene adopts a symmetric scene representation for all agents and lanes (see Fig.~\ref{fig: Symmetric_Scene}).
    }
    \label{fig: dist_vs_topo}
\end{figure}
\section{Introduction}
Trajectory generation is a key component of autonomous driving, encompassing various tasks such as trajectory prediction~\cite{alahi2016social, deo2018convolutional, chai2019multipath, khandelwal2020if, shi2022motion, mo2022multi, gao2020vectornet, liang2020learning, mo2025traffic}, planning~\cite{zeng2019end, mo2023predictive, liu2025hybrid}, and data-driven traffic scene generation~\cite{montali2023waymo, feng2023dense, zhao2024kigras, zhang2026adversarial}. 
Although these tasks differ in their objectives and constraints, they all generate future trajectories conditioned on the surrounding traffic scene, and therefore share a common reliance on accurate, efficient, and scalable traffic scene representation~\cite{ngiam2021scene, jia2023hdgt, shi2024mtr++, mo2025traffic}. 
With the increasing complexity of modern driving environments involving heterogeneous road users, rich map semantics, and intricate spatio-temporal interactions, effectively capturing these factors in a unified and scalable manner has become a fundamental challenge~\cite{nayakanti2023wayformer}.

\moxy{From the perspective of scene representation, existing approaches can be broadly categorized into rasterization-based~\cite{deo2018convolutional, chai2019multipath, cui2019multimodal, zhao2019multi, zeng2019end, salzmann2020trajectron++, suo2021trafficsim, bergamini2021simnet} and vectorization-based~\cite{gao2020vectornet, ngiam2021scene, liang2020learning, shi2022motion, mo2025traffic, mo2023map, mo2023map1} methods. 
Rasterization-based methods suffer from geometric information loss and limited scalability due to fixed-resolution grids.
Vectorization-based approaches represent agents and map elements as vectors and explicitly model their spatial relationships, enabling more compact and structured scene representations~\cite{gao2020vectornet, ngiam2021scene, shi2022motion, mo2025traffic}. 
} 
In recent years, with the release of large-scale autonomous driving datasets~\cite{ettinger2021large, Argoverse} and the emergence of high-performance model architectures such as Transformers~\cite{vaswani2017attention} and graph neural networks (GNNs)~\cite{velickovic2018graph, mo2022multi}, multi-agent trajectory generation has achieved remarkable progress~\cite{gao2020vectornet, liang2020learning, ngiam2021scene, shi2024mtr++, jia2023hdgt}. 
Prediction accuracy on the open benchmarks (e.g., WOMD) continues to improve, with displacement errors steadily decreasing, demonstrating the strong modeling capability of vectorization and deep learning methods in complex traffic scenes. 

In the prevailing Transformer-based approaches, SEPT~\cite{lan2023sept} adopts the most \wzf{straightforward} interaction modeling strategy: it concatenates all $N_a$ agents and $N_l$ lane segments into a single sequence and performs global attention for feature interaction. Its self-attention module computes pairwise attention among all $N_a+N_l$ elements, resulting in a computational complexity of $\mathcal{O}((N_a+N_l)^2)$, which becomes prohibitive in large-scale traffic scenes.
Scene Transformer~\cite{ngiam2021scene} encodes agents and lanes separately, but still applies global attention when modeling agent-to-agent interactions, leading to a complexity of $\mathcal{O}(N_a^2)$. 
To further reduce computational overhead, methods like MTR~\cite{shi2022motion, shi2024mtr++} introduce a distance-based local attention mechanism, restricting lane-to-agent attention to a fixed number of lanes and agent-to-agent attention to a fixed number of nearby agents. 
Although such local-attention designs substantially reduce computational complexity, Transformer-based methods still require predefined limits on the number of input elements (e.g., a fixed maximum number of agents and lane segments), which makes it difficult for them to scale to larger or more complex traffic scenes. 
Moreover, both lane-to-agent and agent-to-agent neighborhoods are constructed purely based on geometric distance.  

Graph-based approaches naturally support variable-sized inputs and therefore provide better scalability to larger traffic scenes. 
Early works such as VectorNet~\cite{gao2020vectornet} \moa{constructs} fully connected graphs between all vectorized elements to ensure rich interaction modeling. 
While expressive, such fully connected designs introduce quadratic complexity with respect to the number of nodes. 
HEAT~\cite{mo2022multi} constructs heterogeneous graphs to model agent interactions, where two agents are connected if their distance is less than a threshold. 
However, distance-driven connections overlook the rich structural priors of roads, \wzf{which strongly constrain and guide driving behavior}. 
LaneGCN~\cite{liang2020learning} constructs a fine-grained lane graph from the HD map and designs a Lane Graph Convolutional Network to capture the complex topology and long-range dependencies of the lane graph. 
This represents an important step toward leveraging HD-map structure in scene representation. 
\moxy{However, the agent-to-agent interaction graph in LaneGCN remains distance-driven, which tends to produce dense and redundant graphs while overlooking topology-induced long-range interactions that are critical for realistic driving behavior, as illustrated in Fig.~\ref{fig: dist_vs_topo}.}

\moxy{Despite the remarkable progress achieved by recent \moa{Transformer and GNN-based approaches}, these methods still suffer from high training and inference costs and significant scalability bottlenecks as the number of agents increases. 
Consequently, designing a compact and expressive representation for large-scale multi-agent traffic scenes is essential for advancing autonomous driving and intelligent transportation systems~\cite{wu2021flow, li2023survey, gao2024evaluation, liu2025gatsim}.} 
We argue that an effective traffic scene representation should 
1) naturally accommodate a variable and potentially large number of traffic elements (e.g., agents and lanes)~\cite{gao2020vectornet}, 
2) integrate HD-map structural priors such as lane topology and connectivity~\cite{liang2020learning}, 
3) model interactions in a manner that captures both spatial geometry and traffic semantics~\cite{shi2024mtr++, mo2025traffic}, and 
4) scale efficiently to large spatial extents~\cite{he2022masked, sun2025sequence, sun2025speed}.
\wzf{Inspired by the observation that meaningful interactions in large-scale traffic scenes are sparse and strongly structured by road topology,} we propose \textbf{SparScene}, a structure-aware traffic scene graph representation designed to capture the sparse yet critical interactions in complex traffic environments. 
SparScene constructs the initial scene graph by directly adopting the HD-map lane topology and connecting each agent to its spatially aligned lanes, ensuring that lane-to-lane edges follow map-defined structure while agent-to-lane edges reflect geometric alignment. 
\kal{Compared to straightforward distance-based interaction construction, SparScene introduces stronger structural constraints by explicitly modeling lane topology and agent–lane associations. 
Although more complex, this design enables topology-guided interaction modeling that avoids redundant connections and focuses on behaviorally critical interactions.}
A hierarchical encoder then models how lane geometry and surrounding agents jointly influence future behavior by performing directed traversal on the lane graph. 
Additionally, SparScene represents all agents and lanes in their ego-centric coordinate systems, forming a \textit{symmetric} scene representation in which every agent perceives the environment from its own viewpoint while preserving consistent relational structure. 
This symmetry is crucial for realistic and unbiased multi-agent modeling and also facilitates scalability to larger areas due to its coordinate-invariant formulation.
\moa{In summary, SparScene introduces a topology-aware sparse scene graph with a symmetric representation, which emphasizes structural relevance over geometric proximity for scalable and efficient multi-agent trajectory generation.}
The contributions of this work can be summarized as: 
\begin{itemize}[leftmargin=10pt]
    \item We propose SparScene, a symmetric and sparse traffic scene graph representation and encoding framework for large-scale trajectory generation. We initialize a heterogeneous agent–lane graph from spatial alignment and unsegmented HD-map lane topology, preserving complete lane connectivity and sparsity, rather than using distance-based heuristics. 
    Based on the graph, we construct a symmetric scene representation and design a three-stage topology-guided encoder, enabling fully parallel, scene-level encoding and multimodal trajectory generation for arbitrary numbers of traffic participants. 
    \item We design a topology-aware strategy for lane-to-agent and agent-to-agent information fusion. Instead of constructing interactions using pre-defined distance thresholds, we leverage lane connectivity and the directional nature of traffic behaviors to determine connections based on directed topology reachability. 
    This produces a compact and interpretable interaction graph that avoids misleading proximity and naturally scales with traffic density. 
    \item We combine SparScene with a lightweight decoder to realize multi-agent multimodal (MAMM) trajectory generation.
    Since the decoder performs no additional interaction modeling, the performance gains directly reflect the effectiveness of the proposed scene representation. On the large-scale WOMD benchmark, SparScene achieves competitive prediction accuracy \wzf{with} extremely low \moa{training and inference costs}, enabling millisecond-level scene-wide trajectory prediction.
\end{itemize}

\section{Related Work}
\label{sec: relatedworks}
\subsection{Scene Graph Construction}
In traffic scenarios, road users and infrastructures exhibit complex spatial, semantic, and temporal relationships. To accurately generate future trajectories, a model must not only capture \moa{each agent’s dynamic states} but also understand its multi-level interactions with surrounding traffic elements.
Therefore, constructing a traffic scene graph has become essential for traffic scene understanding.
Early approaches typically relied on heuristic graph construction methods.
VectorNet~\cite{gao2020vectornet} constructs a fully-connected graph to represent the interactions between road users and map elements. SceneTransformer~\cite{ngiam2021scene} represents a traffic scene with a fixed number of agents and map elements, and then applies Transformers with global attention. This can be regarded as a fully connected graph with a fixed number of nodes. However, such global connections introduce redundant and irrelevant information, increase computational complexity, and may weaken the model’s ability to focus on truly critical interactions.
To address these limitations, several works adopt local connection strategies based on spatial distance, where each agent only interacts with nearby agents or map elements. MTR~\cite{shi2022motion} limits interactions to spatially nearby entities by selecting the 768 nearest map polylines and 16 neighbors around the target agent and applying. Rather than connecting a fixed number of nearest elements,
HiVT~\cite{zhou2022hivt} construct its local graph according to a distance criterion, where each agent is connected to its neighboring agents and lanes, while maintaining a global graph for inter-agent interaction.
While distance-based local strategies improve efficiency, they rely solely on spatial proximity and fail to capture the directional and topological relationships among lanes. 
In LaneGCN~\cite{liang2020learning}, a lane node is defined as the straight line segment formed by any two consecutive points along a lane centerline. Based on this fine-grained definition, LaneGCN constructs a lane graph where nodes represent these short centerline segments and edges capture their topological connectivity. This design allows the model to effectively encode the geometric and directional continuity of the road network. 
Unlike LaneGCN, which builds a fine-grained global lane graph, HGO~\cite{mo2023map, mo2023map1} constructs hierarchical graphs based on candidate centerlines obtained through lane graph search, enabling target-centric map representation and improved computational efficiency.

\moxy{
Although LaneGCN and HGO model lane topology at different levels, the interactions between agents and map elements are still established based on distance heuristics.
In contrast, SparScene constructs interaction graphs by explicitly leveraging lane topological reachability. 
This design imposes strong physical and topological constraints on the interaction graph, leading to a meaningful and sparse scene graph.
}

\subsection{Multi-Agent Trajectory Generation}
Early studies on trajectory generation primarily focused on a single target agent, where the scene context was customized for that agent by normalizing all inputs relative to its position~\cite{deo2018convolutional, liang2020learning, mo2020interaction, mo2023map1, gu2021densetnt, shi2022motion}. However, this agent-centric strategy leads to computational inefficiency when predicting motions for multiple agents~\cite{shi2024mtr++}. 
From the perspective of a single autonomous vehicle, SceneTransformer~\cite{ngiam2021scene} performs multi-agent trajectory prediction by fixing the entire scene’s coordinate system on the ego vehicle and predicting the future trajectories of multiple agents within this ego-centric frame. While this design enables joint prediction from the ego’s viewpoint, the model is sensitive to the choice of coordinate system and struggles to generalize to larger spatial ranges or different reference frames, resulting in degraded performance for off-center agents. 
To overcome the sensitivity to coordinate system selection, HEAT~\cite{mo2022multi} establishes an independent local coordinate frame for each agent and constructs a directed and attributed heterogeneous graph to model the interactions among multiple types of traffic participants. Built upon the graph attention network, HEAT accounts for node heterogeneity and continuous edge features, enabling simultaneous multi-agent trajectory prediction.
HDGT~\cite{jia2023hdgt} further represents map elements and traffic participants in a unified heterogeneous graph, assigns \moa{distinct attention modules~\cite{vaswani2017attention}} to different types of nodes and edges, and encodes spatial relations in each node’s local coordinate frame to enable simultaneous multi-agent trajectory prediction. 
\moxy{MTR++~\cite{shi2024mtr++} extends the MTR~\cite{shi2022motion} framework to symmetric multi-agent motion prediction by removing the target-centric assumption. It models all agents within a unified coordinate system and performs joint decoding in a symmetric manner, and has demonstrated substantially higher efficiency than MTR when generating trajectories for multiple focal agents.
Nevertheless, its scalability is still constrained by fixed-size scene inputs, distance-based interaction modeling, and a relatively heavy decoding architecture.
In contrast, SparScene adopts a fully symmetric design built upon a topology-aware sparse scene graph and employs lightweight decoders for trajectory generation. Our framework imposes no strict limitations on the number of agents or lanes in the scene and enables scalable, efficient, and physically grounded large-scale multi-agent trajectory generation.
}

\section{Method}
\label{sec: method}
\begin{figure*}[ht]
    \centering
    \includegraphics[trim={0cm 0cm 0cm 0cm}, clip, width=1.0\textwidth]{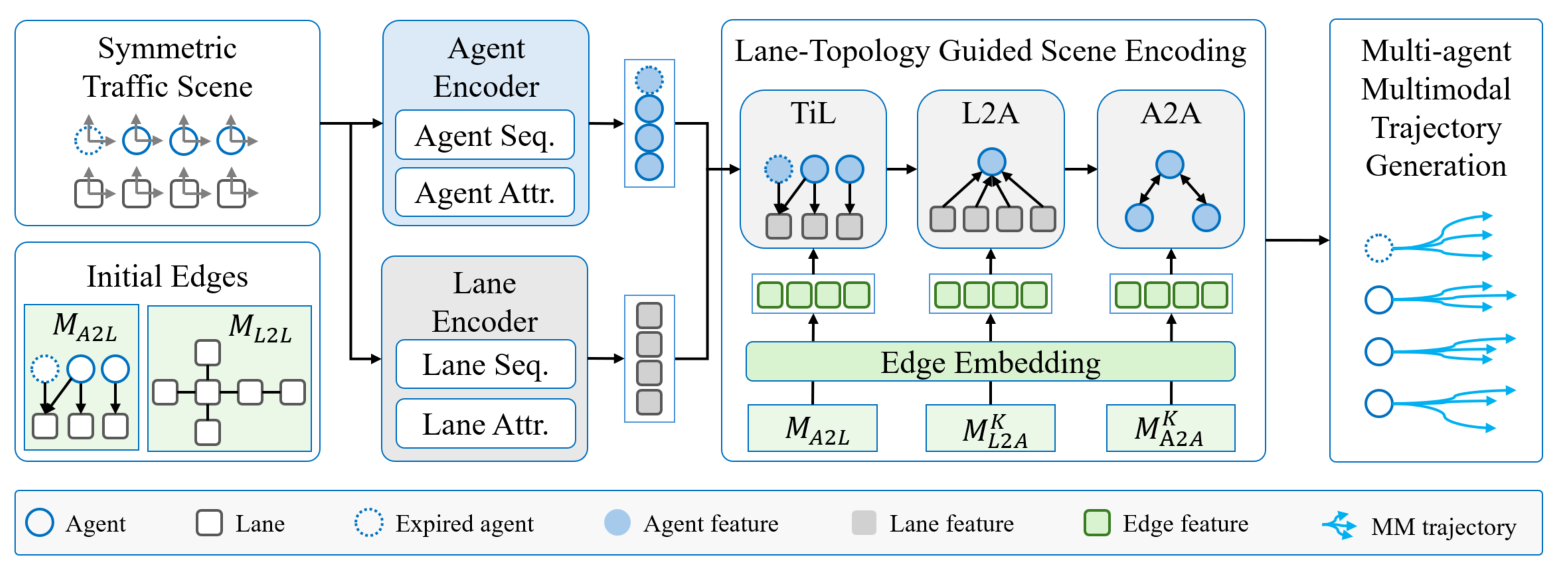}
    \caption{\textbf{Architecture of the SparScene framework.} A symmetric traffic scene is built by representing every agent and lane segment in its own local coordinate frame, ensuring a coordinate-invariant representation. Based on spatial agent–lane alignment and HD-map connectivity, initial agent-to-lane ($M_{A2L}$) and lane–to–lane ($M_{L2L}$) edges are constructed as priors. 
    Agents and lanes are encoded with dynamics and geometric semantics, while edge attributes are embedded into a shared latent space. 
    The encoded nodes are processed by the proposed Lane-Topology Guided Scene Encoding (Sub.Sec.\ref{subsec: scene_enc}), which consists of three stages: 1) Traffic-in-Lane (TiL) aggregates local traffic dynamics into lane nodes; 2) Lane-to-Agent (L2A) propagates traffic-aware lane semantics back to agents; and 3) Agent-to-Agent (A2A) builds sparse, behaviorally feasible interactions induced by lane topology. 
    The resulting representation supports efficient and accurate multi-agent multimodal (MAMM) trajectory prediction with a lightweight decoder.
    }
    \label{fig: SparScene_Framework}
\end{figure*}

\gjt{
Consider a typical traffic scenario populated by heterogeneous agents (e.g., vehicles, pedestrians, and cyclists) and structured map elements (e.g., surface streets, freeways, bike lanes), we refer to scene representation as a latent description of the whole scene that jointly encodes (i) the dynamic states of each agent, (ii) the local traffic, and (iii) the interaction between agents. 
Such a representation can support trajectory forecasting, intention recognition, and other decision-making modules. 
We propose SparScene, focusing on designing a \wzf{highly efficient} traffic scene encoder with \wzf{lane-topology-aware} interaction modeling.
The overall architecture of SparScene is presented in Fig.~\ref{fig: SparScene_Framework}.
}

\begin{figure*}[ht]
    \centering
    \includegraphics[trim={0cm 0cm 0cm 0cm}, clip, width=1.0\textwidth]{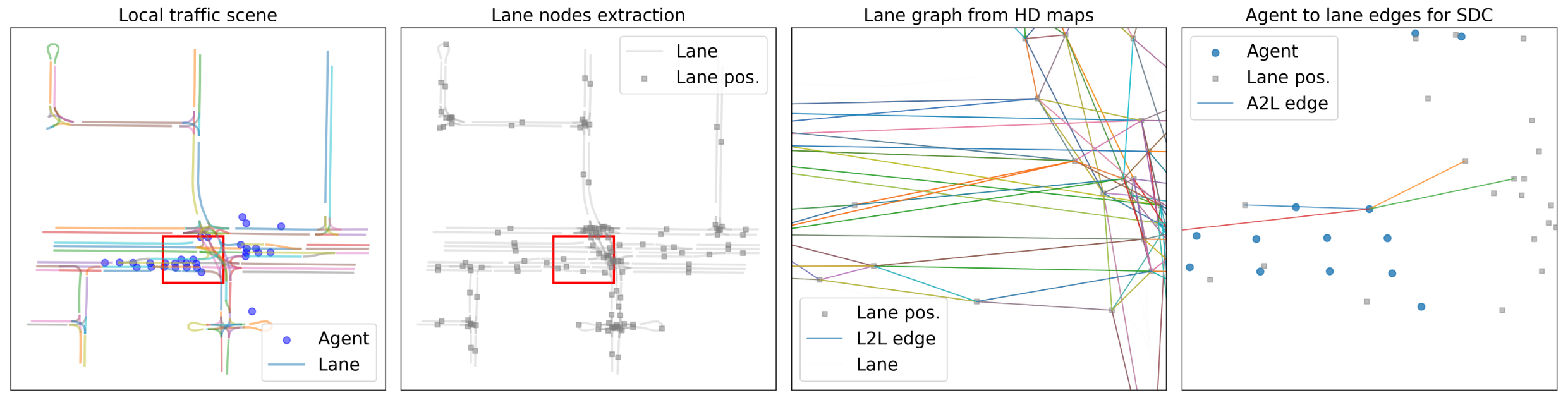}
    \vspace{-6mm}
    \caption{\textbf{Scene graph initialization for SparScene.} 
    The leftmost panel shows a local traffic scene with agents and lane geometries.
    The second and third panels illustrate lane node extraction and lane graph construction from the HD map, respectively.
    The rightmost panel depicts the agent-to-lane interaction edges for the target agent.
    }
    \label{fig: Scene_Graph}
\end{figure*}
\begin{figure}[ht]
    \centering
    \includegraphics[trim={0cm 0cm 0cm 0cm}, clip, width=0.48\textwidth]{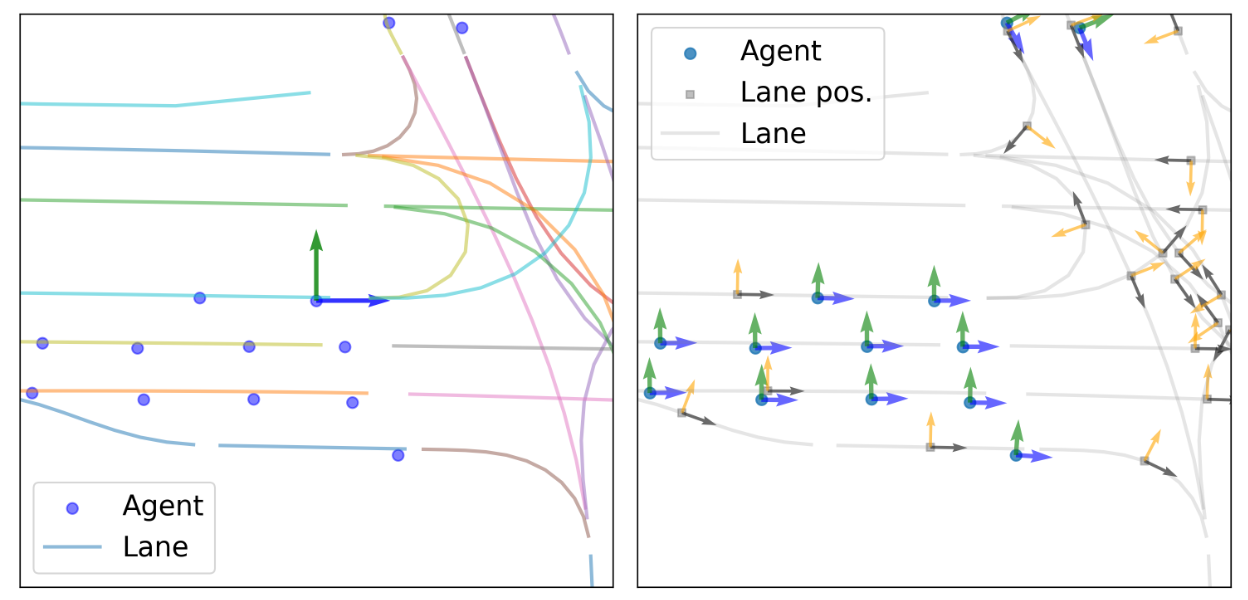}
    \caption{\textbf{Symmetric scene representation.} 
    \textit{Left}: Global scene representation with a fixed origin and orientation.
    \textit{Right}: Symmetric scene representation, where each agent and lane is represented in its own local coordinate system.
    }
    \label{fig: Symmetric_Scene}
\end{figure}
\subsection{Input and Output}
\label{subsec: input_output}
\gjt{We define a traffic scene as a composition of $N_a$ agents and $N_l$ map elements (i.e., lane segments). 
The objective is to infer the future evolution of these agents over a prediction horizon $T_f$ conditional on the observed scene context.}
We address traffic scene representation learning in the task of multi-agent trajectory prediction. 

The historical motion of agent $i \in \{1,\dots,N_a\}$ is represented as a temporal sequence:
$
\mathbf{S}_i = \big[ \mathbf{s}_i^t \big]_{t=-T_h}^{0}, 
\quad 
\mathbf{s}_i^t = (x_i^t, y_i^t, z_i^t, v_{x,i}^t, v_{y,i}^t, \theta_i^t).
$
Positions are converted into displacement vectors.
Each agent additionally carries a static attribute vector with its length ($l_i$), width ($w_i$), height ($h_i$), and type ($c_i$):
$
\mathbf{a}^{\text{agent}}_i = \big(l_i, w_i, h_i, c_i \big).
$
\moa{
The complete agent representation is thus
$
\mathbf{A} = \{\mathbf{A}_1, \dots, \mathbf{A}_{N_a}\}, 
\mathbf{A}_i = [\mathbf{S}_i , \mathbf{a}^{\text{agent}}_i].
$}

The environment of each scenario is represented by $N_l$ lane segments. 
Each lane segment $j \in \{1,\dots, N_l\}$ is described by a sequence of waypoints. 
\gjt{To enable a unified representation of agents' history and lane nodes}, we uniformly resample each lane polyline to a fixed sequence length of $T_h$:
$
\moa{\mathbf{P}_j = \big[ (x_j^k, y_j^k) \big]_{k=1}^{T_h+1}.}
$
Note that the index $k$ denotes spatial ordering on the lane polyline and is not associated with temporal evolution.
In addition to geometric information, lane segments may carry semantic attributes $\mathbf{a}_j^{\text{lane}}$, such as lane type and traffic signals.
The lane representation is
$
\mathbf{L} = \{\mathbf{L}_1, \dots, \mathbf{L}_{N_l}\}, \quad
\mathbf{L}_j = [\mathbf{P}_j, \mathbf{a}_j^{\text{lane}}].
$

With the above definitions, a traffic scenario is represented as a heterogeneous set of nodes composed of agents and lane segments:
$
\mathbf{X} = \{ \mathbf{A}, \mathbf{L} \}.
$
To capture their interactions, we further construct a set of edges $\mathbf{E}$ that represents relations among agents and lane segments. 
Formally, the scene is represented as a heterogeneous graph:
$
\mathbf{G} = \{ \mathbf{X}, \mathbf{E} \}.
$

Given $\mathbf{G}$, the model aims to generate multi-modal future trajectories for multiple agents simultaneously. For each agent $i$, we predict $M$ possible trajectories over a future horizon of length $T_f$:
$
\hat{\mathbf{Y}}_i = \big\{ \hat{\mathbf{y}}_{i}^{(m)} \big\}_{m=1}^{M},
\hat{\mathbf{y}}_{i}^{(m)} \in \mathbb{R}^{T_f \times 2}.
$
The complete model output for a scenario is
$
\hat{\mathbf{Y}} = \{ \hat{\mathbf{Y}}_1, \dots, \hat{\mathbf{Y}}_{N_a} \}.
$

\subsection{Sparse and Symmetric Scenario Graph Initialization}
\label{subsec: scene-enc}
\gjt{To structurally represent the traffic scene without imposing a hard upper bound on the number of elements}, we construct a heterogeneous graph to represent an arbitrary number of agents and lanes, and their connections. 
\moa{The initial graph is sparse}, where lanes are connected as given in the HD maps, and agents are connected to their local lanes only (See Fig.~\ref{fig: Scene_Graph}). 
\gjt{All agents and lanes are represented in their own local coordinate frames, while their global poses are maintained to compute relative distances and heading differences (see Fig.~\ref{fig: Symmetric_Scene}).  } 
The edges also come with types to represent the connection types.

For the lane-to-lane edges, we follow their connections in the lane graph provided by the HD map (see Fig.~\ref{fig: Scene_Graph}). 
\moa{For A$\rightarrow$L edges, we connect an agent to a lane if the distance between them is less than a threshold (e.g., 10 meters). 
For A$\rightarrow$A edges, we leave them blank at the graph construction step and will construct them in the interaction-learning module.}

\subsection{Agent and Lane Encoding}
\label{subsec: agent_lane}
Given the heterogeneous inputs $\mathbf{X} = \{\mathbf{A}, \mathbf{L}\}$, we employ a lightweight \wzf{gated recurrent unit (GRU)~\cite{chung2014empirical}} encoder to encode each agent’s historical states ($\mathbf{S}_i$) and capture temporal dependencies. The resulting motion features are fused with high-dimensional representations of agent type and geometric attributes ($\mathbf{a}^{\text{agent}}_i$) through an MLP, producing each agent's dynamics embedding $\mathbf{h}_i^{\text{agent}}$ that serves as input for subsequent interaction modeling. 
Similarly, we employ a GRU-based lane encoder to model the spatial continuity and directional dependencies of lane centerlines represented by waypoint sequences ($\mathbf{P}_j$). The GRU captures the geometric continuity and directional dependencies along the lane, while lane attribute ($\mathbf{a}_j^{\text{lane}}$) embeddings are fused through an MLP to obtain the lane representation $\mathbf{h}_j^{\text{lane}}$.
The encoded features serve as the initial node representations of the scene graph:
$
\mathbf{H}^{(0)} = \big[ \mathbf{h}_1^{\text{agent}},\dots,\mathbf{h}_{N_a}^{\text{agent}}, 
\mathbf{h}_1^{\text{lane}},\dots,\mathbf{h}_{N_l}^{\text{lane}} \big].
$

\subsection{Edge Encoding}
\label{subsec: edge_enc}
The scene graph consists of agent and lane nodes, encoded by the modules described in Sub.Sec.~\ref{subsec: agent_lane}. 
To model relational dependencies between nodes, we construct a unified geometric embedding for each directed edge $j \!\rightarrow\! i$.
\gjt{Let node $i$ and node $j$ have global poses $(\mathbf{p}_i, \theta_i)$ and $(\mathbf{p}_j, \theta_j)$, \wzf{respectively,} where $\mathbf{p}_i, \mathbf{p}_j \in \mathbb{R}^2$ denote positions and $\theta_i, \theta_j$ denote headings. For an edge from source node $j$ to target node $i$, we first express the relative position and heading in the local coordinate frame of node $i$:
$
\Delta \mathbf{p}_{ij} = R(-\theta_i)\,(\mathbf{p}_j - \mathbf{p}_i)
= [\Delta x_{ij}, \Delta y_{ij}]^\top,
$
$
\Delta \theta_{ij} = \mathrm{wrap}(\theta_j - \theta_i).
$
where $R(\cdot)$ is the $2{\times}2$ rotation matrix and $\mathrm{wrap}(\cdot)$ maps angles to $(-\pi,\pi]$, 
}
This ego-centric formulation enables each node to perceive its neighbors from its own perspective, \wzf{yielding a coordinate-invariant (symmetric) relational representation.}
To ensure directional continuity, the orientation term is encoded as $(\cos(\Delta \theta_{ij}), \sin(\Delta \theta_{ij}))$.
In addition, each edge carries a semantic type embedding to distinguish heterogeneous interactions (e.g., agent-agent interaction, agent-lane alignment, lane-lane connectivity). 
The final edge representation is obtained by concatenating the geometric descriptor
$
[\Delta x_{ij}, \Delta y_{ij}, \cos(\Delta \theta_{ij}), \sin(\Delta \theta_{ij})]
$
with the type embedding and projecting it into a latent space via a lightweight MLP. 
These edge embeddings $\mathbf{e}_{ij}$ serve as relational features to support reasoning in the scene graph.

\subsection{Lane-Topology Guided Scene Encoding}
\label{subsec: scene_enc}
In real-world traffic, interactions are constrained by road topology, lane structure, and traffic rules rather than spatial proximity alone. The lane graph offers a natural structural prior for modeling such interactions, yet distance-based graph construction often overlooks it. 
We therefore incorporate this prior and propose a lane-topology guided scene encoding framework with a three-stage interaction modeling process: 1) dynamic agent states are first sent into their aligned local lanes; 2) lane semantics, augmented through topology-aware graph search (capturing upstream/downstream connectivity and potential maneuver relations), are propagated back to agent nodes; and 3) lane topology further guides the construction of A$\rightarrow$A edges by linking agents on the same or topologically connected lanes, forming a sparse yet meaningful interaction graph. 
This lane-guided structural encoding naturally inherits the sparsity of road topology, offering both interpretability and computational scalability for large-scale multimodal trajectory generation.

\subsubsection{Traffic in Lane (TiL)}
\label{subsubsec: a2l}
This stage introduces local traffic dynamics into lane nodes, transforming lane representations, which originally encode geometric and signal semantics, into traffic-aware semantic nodes.
Specifically, we construct agent-to-lane edges based on pure spatial alignment between agents and lanes, with no lane-topology search. We employ the message passing mechanism to aggregate agent motion embeddings into their corresponding lane nodes. 
By summarizing local traffic dynamics onto lane nodes, TiL establishes lanes as a compact and informative semantic carrier for the subsequent lane-to-agent information fusion stage. 
Once their motion dynamics have been summarized to lane nodes, currently invisible agents no longer need to be explicitly modeled, as their effects are already preserved in the TiL representation.

\moxy{
\textbf{Unified Message Passing}\wzf{:}
We adopt a unified Message Passing scheme to model interactions in traffic scenes. 
The feature of node $i$ at layer $k$ is updated as
\begin{equation}
\begin{aligned}
\mathbf{x}_i^{(k)} &=
\text{UPDATE}\Big(
\mathbf{x}_i^{(k-1)}, \\
&\quad \text{AGG}\big(
\{ \phi(\mathbf{x}_i^{(k-1)}, \mathbf{x}_j^{(k-1)}, \mathbf{e}_{ij})
\mid j \in \mathcal{N}(i) \}
\big)
\Big),
\end{aligned}
\end{equation}
where $\mathcal{N}(i)$ denotes the neighbor set of node $i$, 
$\phi(\cdot)$ is a learnable message function, 
AGG$(\cdot)$ is the message aggregation operator, and UPDATE$(\cdot)$ is the node update function.
We implement AGG using the multi-head attention mechanism~\cite{vaswani2017attention, velickovic2018graph}. Moreover, a gated update mechanism is adopted in UPDATE to fuse past node states and aggregated interaction messages:
$
\mathbf{x}_i^{(k)} = \mathbf{x}_i^{(k-1)} + \mathbf{g}_i \odot \big( \mathbf{m}_i - \mathbf{x}_i^{(k-1)} \big),
$
where $\mathbf{m}_i$ denotes the aggregated message from neighbors, 
$\mathbf{g}_i$ is the gate coefficient, and
$\odot$ is the element-wise product.
Based on this unified message passing operator, we further construct the following interaction modules.
}

\subsubsection{Lane to Agent (L2A)}
\label{subsubsec: l2a}
Based on the traffic-aware lane nodes obtained from the TiL stage, the goal of L2A is to propagate lane-level semantics back to currently visible agents, enabling each agent to acquire structured contextual information that is relevant to its potential future maneuvers.
We construct semantically meaningful L$\rightarrow$A edges through directional lane-topology expansion, which consists of two complementary search operations:
\wzf{a)} The \textit{Omnidirectional Expansion} \wzf{(OE)}  operation expands one step along all lane-connectivity directions from each lane node, thereby incorporating neighboring lane nodes that may influence short-term vehicle behavior. It enables the modeling of potential local interactions, such as lane-changing competition, which can impact a vehicle’s immediate motion trend, providing essential local interaction semantics for trajectory prediction.
\wzf{b)} The \textit{Forward Expansion} \wzf{(FE)} takes the directional nature of traffic behaviors into account and expands only one step along feasible forward-driving directions. This forward-only propagation naturally filters out lane nodes that are irrelevant to the vehicle’s future trajectory. By restricting propagation to only maneuver-feasible directions, FE not only reduces semantic noise but also eliminates unnecessary computational overhead.
We integrate OE and FE to construct the final L$\rightarrow$A connectivity for message passing, expanding along the lane graph for up to $K$ topological steps:
\begin{equation}
    M^{K}_{L2A} = \bigcup_{k=0}^{K} M_{L2L}^{(k)}  \wzf{\cdot}\, M_{L2A},
    \label{eq: M_L2A}
\end{equation}
where $M_{L2L}^{(0)} = I$, and each $M_{L2L}^{(k)}$ $(k \ge 1)$ denotes the product of $k$ lane-to-lane adjacency matrices selected from $M_{L2L-O}$ (Omnidirectional) and $M_{L2L-F}$ (Forward). The union $\cup$ denotes element-wise logical OR (edge-set union) over adjacency matrices. 
In practice, we adopt a hybrid expansion sequence (e.g., O$\rightarrow$F$\rightarrow$F), resulting in sparse yet behaviorally meaningful L$\rightarrow$A edges for trajectory prediction.

\subsubsection{Agent to Agent (A2A) Interaction}
\label{subsubsec: a2a}
After the TiL and L2A stages, each agent has already acquired awareness of \moa{local} road structures and traffic flow. The goal of A2A is to further model high-level interactions between agents based on this information. 
Unlike distance-based interaction graph construction, we argue that potential interactions between traffic participants are not solely determined by spatial proximity but also constrained by the road structure. 
Two agents are connected only if their associated lanes are topologically reachable within a limited number of hops, implying potential interactions. These lane-guided edges are formulated as:
\begin{equation}
    M_{A2A}^{K} = \bigcup_{k=0}^{K} M_{A2L} \cdot M_{L2L}^{(k)} \cdot M_{L2A}, 
    \label{eq: M_A2A}
\end{equation}
\moa{
where $M_{L2A}=M^{\top}_{A2L}$ represents the directed edges from lanes to agents.} 
$M_{L2L}^{(k)}$ and $\cup$ are defined as in Eq.~\ref{eq: M_L2A}.
Note that when $K=0$, $M_{A2A}^{K}$ reduces to connecting only agents aligned to the same lane.
This interaction graph construction avoids both false-positive interactions caused by geometric closeness (e.g., vehicles separated by a physical median barrier) and false-negative cases where \wzf{critical-conflict} vehicles have not yet approached spatially but are topologically connected (e.g., imminent merging).

\subsection{Multi-Agent Trajectory Generation}
\label{subsec: traj_gen}
To fairly assess the capability of our scene representation, we evaluate it on the multi-agent multimodal trajectory prediction task and employ a lightweight decoding module to avoid the influence of complex decoders. 
\gjt{The goal of this decoder is not to compete with sophisticated forecasting heads, but to serve as a simple readout that probes how well the encoded scene features support future-motion prediction. The decoder does not perform any additional interaction modeling. Instead, it directly maps the encoded scene features to future motion through an MLP-based regression head.}
Each agent node is represented by three feature components provided by the scene encoder: 1) dynamic features, 2) local lane features, and 3) interaction features. 
These features are concatenated as decoder inputs.
Considering the behavioral differences across agent categories (e.g., vehicles, pedestrians, cyclists), we design separate decoders for each type. 
\gjt{To capture multimodality, we further maintain, for each category, a small set of learnable mode embeddings that represent high-level behavioral intents.} 
Agents of the same category share the same mode embeddings and a common decoder, enabling behavior-specific modeling without sacrificing efficiency.  
The overall output for a scene with $N_a$ target agents is 
$\hat{\mathbf{Y}}  \in \mathbb{R}^{N_a \times M \times T_f \times 2}$, \gjt{which is supervised during training. In this way, trajectory prediction acts as a downstream probe: performance improvements can be primarily attributed to a stronger scene representation, rather than to decoder complexity.}
By leveraging symmetric scene representations and shared type-specific decoders, the proposed architecture supports efficient parallel prediction over all agents in the scene.

\subsection{Training Objective}
\label{subsec: MTP_loss}
Focusing on trajectory prediction accuracy, we employ only the regression component of the MTP Loss~\cite{cui2019multimodal} and do not perform any mode probability prediction. 
For a target agent $i$ with ground-truth future trajectory $\mathbf{Y}_i \in \mathbb{R}^{T_f \times 2}$, the decoder outputs $M$ trajectory hypotheses $\{\hat{\mathbf{y}}_i^{(m)}\}_{m=1}^{M}$. 
We apply a best-of-$M$ supervision strategy, where the regression error is computed using the Smooth L1 (Huber) loss as:
\begin{equation}
\mathcal{L}_i = \min_{m \in \{1,\dots,M\}} \text{SmoothL1}\left(\hat{\mathbf{y}}_i^{(m)}, \mathbf{Y}_i\right).
\end{equation}
For a batch of scenarios, the overall objective is averaged over all $N_a$ agents:
$
\mathcal{L} = \frac{1}{N_a} \sum_{i=1}^{N_a} \mathcal{L}_i.
$

\begin{table*}[t]
\caption{Multimodal trajectory prediction performance on the WOMD test set}
\vspace{-2mm}
\centering
\begin{tabular}{l|r|c|c|c } 
\toprule
Method & Reference & minADE (m) $\downarrow$ & minFDE (m) $\downarrow$ & MR $\downarrow$ \\
\midrule
\texttt{LSTM}~\cite{ettinger2021large} & - & 1.0065 & 2.3553 & 0.3750 \\
\texttt{ReCoAt}~\cite{huang2022recoat} & ITSC 2022 & 0.7703 & 1.6668 & 0.2437 \\
\texttt{DenseTNT}~\cite{gu2021densetnt} & ICCV 2021 & 1.0387 & 1.5514 & 0.1779 \\
\texttt{MotionCNN}~\cite{konev2022motioncnn} & CVPRw 2021 & 0.7400 & 1.4936 & 0.2091 \\
\texttt{SceneTransformer}~\cite{ngiam2021scene} & ICLR 2022 & 0.6117 & 1.2116 & 0.1564 \\
\texttt{LaneGCN}~\cite{liang2020learning} & ECCV2020 & 0.7632 & 1.5589 & 0.2293 \\
\texttt{MTR}~\cite{shi2022motion} & NeurIPS 2022 & 0.6050 & 1.2207 & 0.1351 \\
\texttt{SceneGNN}~\cite{mo2025traffic} & T-ITS 2025 & 0.6029  & 1.2176 & 0.1629 \\
\texttt{HDGT}~\cite{jia2023hdgt} & T-PAMI 2023 & 0.5933 & 1.2055 & 0.1511 \\
\texttt{MTR++}~\cite{shi2024mtr++} & T-PAMI 2024  & 0.5906 & 1.1939 & 0.1298 \\
\midrule
\texttt{SparScene (Ours)} & - & 0.5924 & 1.1834 & 0.1559 \\
\bottomrule
\end{tabular}
\label{tab: womd_test}
\end{table*}
\begin{table*}[t]
\centering
\begin{threeparttable}
\caption{Comparison of model efficiency and prediction performance on the WOMD test set}
\label{tab:efficiency_comparison}
\begin{tabular}{lrrrcccr}
\toprule
Method & Reference & \#Params & Latency $\downarrow$ & minADE (m) $\downarrow$ & minFDE (m) $\downarrow$ & MR $\downarrow$ & Device \\
\midrule
\texttt{SceneTransformer}~\cite{ngiam2021scene} & ICLR 2022 & 15.3 M &  52 ms   & 0.6117 & 1.2116 & 0.1564 & V100 \\
\texttt{HDGT}~\cite{jia2023hdgt}                & T-PAMI 2023 & 12.1 M &  1320 ms & 0.5933 & 1.2055 & 0.1511 & 8000  \\
\texttt{MTR}~\cite{shi2022motion}               & NeurIPS 2022 & 65.8 M &  193 ms & 0.6050 & 1.2207 & 0.1351 & 8000  \\
\texttt{MTR++}~\cite{shi2024mtr++}              & T-PAMI 2024 & 86.6 M &  118  ms & 0.5906 & 1.1939 & 0.1298 & 8000  \\
\midrule
\texttt{SparScene (Ours)} & - & 3.2 M  &  5 ms  & 0.5924 & 1.1834 & 0.1559  & 5090 \\
\bottomrule
\end{tabular}
\begin{tablenotes}
\footnotesize
\item Latency is measured by predicting 32 agents per scenario. 
SceneTransformer results are reported on an NVIDIA V100 GPU, 
MTR, MTR++, and HDGT on an NVIDIA Quadro RTX 8000 GPU, and our SparScene on an NVIDIA RTX 5090 GPU for all agents in a scenario. 
Parameter counts and metrics of baselines are quoted from MTR++~\cite{shi2024mtr++}.
\end{tablenotes}
\end{threeparttable}
\end{table*}

\section{Experiments}
\label{subsec: exps}
\subsection{Experimental Setup}
\paragraph{Dataset and Task}
We conduct experiments on the Waymo Open Motion Dataset (WOMD)~\cite{ettinger2021large}, one of the largest and most diverse motion forecasting datasets for autonomous driving. WOMD contains more than 570 hours of real-world driving data collected across six U.S. cities at 10 Hz, comprising over 100,000 traffic scenarios, each recorded as a continuous 20-second time sequence. The dataset includes multiple types of traffic participants, such as vehicles, pedestrians, and cyclists, all labeled using a high-precision 3D auto-labeling system, and each scenario is accompanied by high-definition (HD) maps. Due to its rich interactive driving behaviors, WOMD is well-suited for traffic scene representation-related tasks such as multi-agent trajectory prediction. 

For trajectory prediction, WOMD officially provides 9-second prediction segments, including 1 second (11 frames) of observed history and 8 seconds (80 frames) of future annotations, resulting in over 500,000 highly interactive prediction instances. We follow the official train/validation/test splits. The test set only provides maps and historical trajectories, while ground-truth future trajectories are withheld, requiring predictions to be submitted to the official WOMD server for evaluation. Each \moa{scenario} contains up to eight target agents, for which multimodal future trajectories must be predicted.

\paragraph{Evaluation Metrics} 
This work aims to develop an efficient traffic scene representation rather than optimizing a specific trajectory decoder. We choose multi-agent multimodal trajectory prediction as our downstream evaluation task. 
\moa{Trajectory forecasting has recently received extensive attention in autonomous driving research, and its performance directly reflects the quality of scene representation.} 
This task simultaneously examines whether the representation captures spatiotemporal dependencies among heterogeneous elements, and whether it supports future uncertainty (multimodality) as well as multi-agent interactions (multi-agent reasoning). In addition, our method targets efficient modeling in large-scale scenarios; therefore, predicting multiple agents in parallel also highlights the benefit of our representation in terms of both accuracy and efficiency.
\moa{We adopt the most widely used and intuitive forecasting metrics: minimum average displacement error (minADE), minimum final displacement error (minFDE), and miss rate (MR) as defined in WOMD~\cite{ettinger2021large}.} These metrics directly reflect how well a scene representation supports downstream prediction quality.

\paragraph{Implementation Details} 
The proposed model is implemented using PyTorch~\cite{NEURIPS2019_9015} and PyTorch Geometric (PyG)~\cite{Fey/Lenssen/2019}.
\textbf{For benchmarking}, we train one model for the results in Tab.~\ref{tab: womd_test}, Tab.~\ref{tab:efficiency_comparison}, Tab.~\ref{tab:exp_scalability_mtr},
and Tab.~\ref{tab:scalability_5000}.
The model for benchmarking is trained from scratch on the entire WOMD training set using a single GPU (NVIDIA RTX 5090), within 21 hours. 
The model is trained for 64 epochs with a batch size of 64.
Based on the OFF strategy, we search the lane graph to generate \moa{L$\rightarrow$A and A$\rightarrow$A edges}. The information fusion module of SparScene is built using attention-based GNNs, where we use a 1-layer GNN for TiL, and stack 2-layer GNNs for both L2A and A2A modules. 
For each scene, the model outputs six future trajectories of 80 steps for every agent, forming the multimodal trajectory predictions. 
We use AdamW as the optimizer, with an initial learning rate of 3e-4 and a weight decay of 1e-4. A cosine annealing schedule is applied during training, which gradually decays the learning rate to approximately 1e-6 at the end of training.
\textbf{For investigation}, we train the models on 20\% of the training data for 32 epochs with a batch size of 32, and validate them on more than $41,000$ scenarios from the validation set. 
All other \moa{training} settings are the same as in benchmarking.

\newcolumntype{C}[1]{>{\centering\arraybackslash}m{#1}}
\begin{table*}[t]
\centering
\begin{threeparttable}
\caption{Scaling efficiency comparison with MTR and MTR++ on WOMD validation set}
\label{tab:exp_scalability_mtr}
\begin{tabular}{lc|C{1.2cm}C{1.2cm}C{1.2cm}|C{1.2cm}C{1.2cm}C{1.2cm}}
\toprule
\multirow{3}{*}{Method} & \multirow{3}{*}{minFDE $\downarrow$} & \multicolumn{6}{c}{Efficiency with Increasing Numbers of Focal Agents} \\
&
& \multicolumn{3}{c|}{Inference Latency $\downarrow$ (ms)} & \multicolumn{3}{c}{Memory Cost $\downarrow$ (GB)} \\
&  & 8 & 16 & 32 & 8 & 16 & 32 \\
\midrule
\texttt{MTR}~\cite{shi2022motion}    & 1.2207 & 84 & 123 & 193 & 5.2 & 7.1 & 15.6 \\
\texttt{MTR+}~\cite{shi2024mtr++}    &  - & 67 & 78 & 98 & 2.9 & 3.2 & 4.7 \\
\texttt{MTR++}~\cite{shi2024mtr++}   & 1.1939 & 77 & 90 & 118 & 3.1 & 3.4 & 5.2 \\
\midrule
\texttt{SparScene (Ours)}    & 1.1834 &  5 & 5 & 5 & 0.75 & 0.75 & 0.75 \\
\bottomrule
\end{tabular}
\begin{tablenotes}
\footnotesize
\item Performance metrics are reported on WOMD following the official evaluation protocol.
Inference latency and memory are measured by predicting multiple focal agents per scenario; we report the average cost for \{8, 16, 32\} focal agents.
Baseline numbers are quoted from MTR++~\cite{shi2024mtr++} (hardware as reported therein). Our measurements are obtained on RTX~5090 unless otherwise specified. 
Our results are obtained by running on a scenario with 91 agents and 458 lanes. The model directly outputs all agents' future trajectories. 
\end{tablenotes}
\end{threeparttable}
\end{table*}

\subsection{Prediction Results on the WOMD Benchmark}
To comprehensively evaluate the performance of SparScene, we compare it with three representative categories of trajectory generation methods on the WOMD test set, including mainstream models based on CNNs, GNNs, and Transformers.
Specifically, CNN-based methods (e.g., MotionCNN~\cite{konev2022motioncnn}) capture local spatio-temporal features through convolutional structures;
GNN-based methods (e.g., DenseTNT~\cite{gu2021densetnt}, LaneGCN~\cite{liang2020learning}, HDGT~\cite{jia2023hdgt}, and SceneGNN~\cite{mo2025traffic}) model the interactions among multiple agents using graph structures;
and Transformer-based methods (e.g., SceneTransformer~\cite{ngiam2021scene} and MTR++~\cite{shi2024mtr++}) perform end-to-end scene modeling with local attention mechanisms.
SparScene mainly targets scene representation, and therefore, we compare it with the above models only in terms of displacement-based trajectory prediction metrics. This evaluation isolates the contribution of scene representation from other task-specific components such as multimodality modeling or decoding strategies.

Tab.~\ref{tab: womd_test} presents the prediction results of SparScene and three representative categories of baseline methods on the WOMD test set.
As shown, SparScene achieves outstanding performance across all displacement-based metrics, including minADE, minFDE, and MR.
It outperforms CNN, GNN, and Transformer-based approaches in prediction accuracy, demonstrating the overall effectiveness of the proposed scene representation method for multi-agent trajectory prediction.

Beyond accuracy, as a sparse and lightweight scene representation, SparScene delivers these competitive results with significantly lower computational overhead. We next analyze its efficiency and scalability in large-scale multi-agent forecasting.

\subsection{Efficiency Comparison with Baselines}

\subsubsection{Overall Efficiency}
To evaluate model efficiency, we compare SparScene with several representative baselines under the same evaluation protocol.
Tab.~\ref{tab:efficiency_comparison} summarizes the number of parameters, inference latency, and prediction performance on the WOMD test set.
For a fair comparison, the results of baseline methods are directly copied from their original papers, while SparScene is evaluated on an NVIDIA RTX 5090 GPU.
As shown, SparScene requires only about 3.2M parameters and 5 ms latency to achieve prediction accuracy comparable to strong Transformer-based models such as MTR++, demonstrating the superior efficiency and lightweight nature of our sparse scene representation. 
Compared with the strongest baseline MTR++, the number of parameters is reduced by approximately 96\%, and the inference speed is improved by about 23 times.
These results highlight that the proposed sparse graph learning mechanism effectively reduces model complexity and accelerates inference, providing a lightweight and scalable solution for large-scale scene representation.

\subsubsection{Scaling Efficiency}
\moa{
Recent advances in the WOMD leaderboard highlight the importance of shared scene encoding for multi-agent forecasting as demonstrated in MTR++~\cite{shi2024mtr++}. MTR+ is a variant of MTR++ that improves over the original MTR~\cite{shi2022motion} by introducing a symmetric scene context encoder that allows multiple agents to share encoded context features, thereby avoiding repeated scene processing.} This design substantially reduces latency and memory usage when the number of agents increases, while maintaining strong forecasting accuracy. As reported in~\cite{shi2024mtr++} and Tab.~\ref{tab:exp_scalability_mtr}, the symmetric encoder reduces latency from 193 ms to 98 ms and memory from 15.6 GB to 4.7 GB when expanding prediction from 8 to 32 agents, demonstrating the benefit of shared context modeling.

Results in Tab.~\ref{tab:exp_scalability_mtr} show that as the number of predicted agents grows from 8 to 32, MTR, MTR+, and MTR++ exhibit nearly linear growth in inference latency and memory consumption, whereas the inference time of SparScene remains almost constant at around 5 ms and 0.75 GB memory within this range.
This advantage comes from the scene-level parallel \moa{generation} mechanism, where SparScene performs a single forward pass over the entire traffic scene graph to predict the future trajectories of all agents.
Overall, SparScene demonstrates remarkable scalability in a typical traffic scene involving 91 agents and 458 lanes.

\subsection{Scaling to Regional-level Trajectory Generation}
Existing scene-level trajectory generation methods typically operate on a single road segment or intersection, whereas modern traffic systems require trajectory generation across multiple intersections, or even large-scale road networks. Therefore, we evaluate the scalability of SparScene on a larger regional traffic area. Because WOMD does not contain a single scenario with more than one thousand traffic participants, we construct larger regional inputs by combining multiple independent scenes to simulate a real-world road network.

\textbf{Enlarged scene via graph batching} is realized by the advanced mini-batching in PyG~\cite{Fey/Lenssen/2019}. 
Unlike common deep learning models, where batch introduces an additional dimension, and each sample is processed independently, SparScene merges multiple scene graphs into one enlarged graph by concatenating all agent and lane nodes and stacking their adjacency matrices along the block diagonal. 
\begin{table}[t]
\centering
\begin{threeparttable}
\caption{Inference scalability of SparScene}
\label{tab:scalability_5000}
\begin{tabular}{lccccc}
\toprule
\# Agents   & 91 & 143 & 228 & 2,199 & 5,620 \\
\# Lanes    & 458 & 883 & 1,303 & 6,063 & 17,463 \\
\# Scenes   & 1 & 2 & 4 & 32 & 80 \\
\midrule
Inference Latency (ms) & 5 & 5 & 5 & 20 & 54 \\
GPU Memory Cost (GB)   & 0.75 & 0.81 & 0.85 & 1.5 & 2.9 \\
\bottomrule
\end{tabular}
\begin{tablenotes}
\footnotesize
\item Latency values are averaged over 1000 forward passes after 100 warm-up runs, using the same batched graph as input on a single NVIDIA RTX 5090 GPU. 
\end{tablenotes}
\end{threeparttable}
\end{table}

\wzf{Tab.~\ref{tab:scalability_5000} reports the scalability of SparScene as the traffic scene size increases.} The first column (91 agents, 458 lanes) is a standard WOMD scene sampled with batch size 1. The larger cases do not naturally exist in the dataset; instead, \moa{they are constructed by mini-batching multiple WOMD scenes, respectively, resulting in enlarged regional-scale scene graphs with up to 5,620 agents.} 
Because of the advanced mini-batching mechanism~\cite{Fey/Lenssen/2019}, SparScene performs full-scene trajectory generation in one pass. It can be observed that the inference latency only grows with the total graph size and remains as low as 20 ms and 54 ms even for 2,199 and 5,620 agents, respectively. Meanwhile, the GPU memory cost of SparScene scales sub-linearly with scene size, requiring only 2.9 GB even for a regional-scale input containing 5,620 agents and 17,463 lanes.
Notably, SparScene produces multimodal trajectories for every agent simultaneously, yielding a full-scene output of shape [5620, 6, 80, 2].
These results demonstrate that SparScene can scale beyond scenario-level trajectory generation and supports real-time regional-level generation under heterogeneous traffic conditions.

It is worth noting that all latency and memory results are obtained using the exact model we used for submission to the WOMD test server, where SparScene achieves competitive prediction accuracy as shown in Tab.~\ref{tab: womd_test}. The efficiency does not rely on reduced model capacity, lightweight approximation, or inference-specific optimization, demonstrating that SparScene offers both high accuracy and \wzf{strong} scalability.

\begin{table}[t]
\centering
\begin{threeparttable}
\caption{Ablation study on SparScene}
\label{tab: ablation_1}
\setlength{\tabcolsep}{6pt}
\renewcommand{\arraystretch}{1.1}
\begin{tabular}{l|ccc|c}
\toprule
Variant & TiL & L2A & A2A & minFDE@8s (m) $\downarrow$  \\
\midrule
\texttt{AL+G2}   &  & \ding{51} &  & 3.28   \\
\texttt{AL+G12}  & \ding{51} & \ding{51} & & 2.89  \\
\midrule
\texttt{SparScene} & \ding{51} & \ding{51} & \ding{51} & 2.73  \\
\bottomrule
\end{tabular}
\begin{tablenotes}
\footnotesize
\item Trained on 20\% training set and validated on more than 41,000 scenarios in the validation set.
\end{tablenotes}
\end{threeparttable}
\end{table}

\subsection{Ablation Study}
We conduct a series of ablation studies to evaluate the effectiveness of each component and design choice in SparScene.
Following the ablation protocol in MTR++~\cite{shi2024mtr++}, we sample 20\% of the WOMD training set due to the large scale of the full dataset.
On this reduced dataset, all ablation variants are trained for fewer epochs with a smaller batch size, while keeping all other training configurations identical to the benchmarking model.
For evaluation, we report minADE and minFDE at the longest prediction horizon of 8 seconds.
This differs from the benchmarking setting, where metrics are averaged across the three horizons (3 s, 5 s, and 8 s).

\subsubsection{Effects of Each Module} 
\label{subsubsec: ablation_1}
\moa{
Tab.~\ref{tab: ablation_1} presents the ablation results of the three scene encoding modules in SparScene, namely TiL (\texttt{G1}~\ref{subsubsec: a2l}), L2A (\texttt{G2}~\ref{subsubsec: l2a}), and A2A (\texttt{G3}~\ref{subsubsec: a2a}).}
We begin with the \texttt{AL+G2} configuration, where the target agent’s historical dynamics (\texttt{A}) and its associated lane features  (\texttt{L}) are encoded for trajectory generation.
The L2A module (\texttt{AL+G2}) constrains agent motion using lane features propagated through the \texttt{OFF}-based lane graph exploration, highlighting the importance of road-structure guidance in future prediction.
Adding the TiL module (\texttt{AL+G12}) further improves performance by explicitly injecting traffic semantics into each lane node, enabling it to represent the dynamic traffic state within the lane. Importantly, these semantics include not only currently visible agents but also those that appeared in the same lane during the historical horizon but are no longer visible in the current frame. This provides each lane node with a more complete spatio-temporal context, allowing the model to focus on subsequent computations on currently visible agents while reducing redundant message passing.
Finally, incorporating the A2A module into \texttt{AL+G12} yields the full SparScene model, which significantly strengthens multi-agent interaction modeling and leads to reduced displacement error.
Overall, this ablation study demonstrates that the hierarchical interaction modeling strategy of SparScene effectively enhances scene understanding and improves trajectory generation accuracy.

\begin{figure}[t]
    \centering
    \includegraphics[trim={0cm 0cm 0cm 0cm}, clip, width=0.48\textwidth]{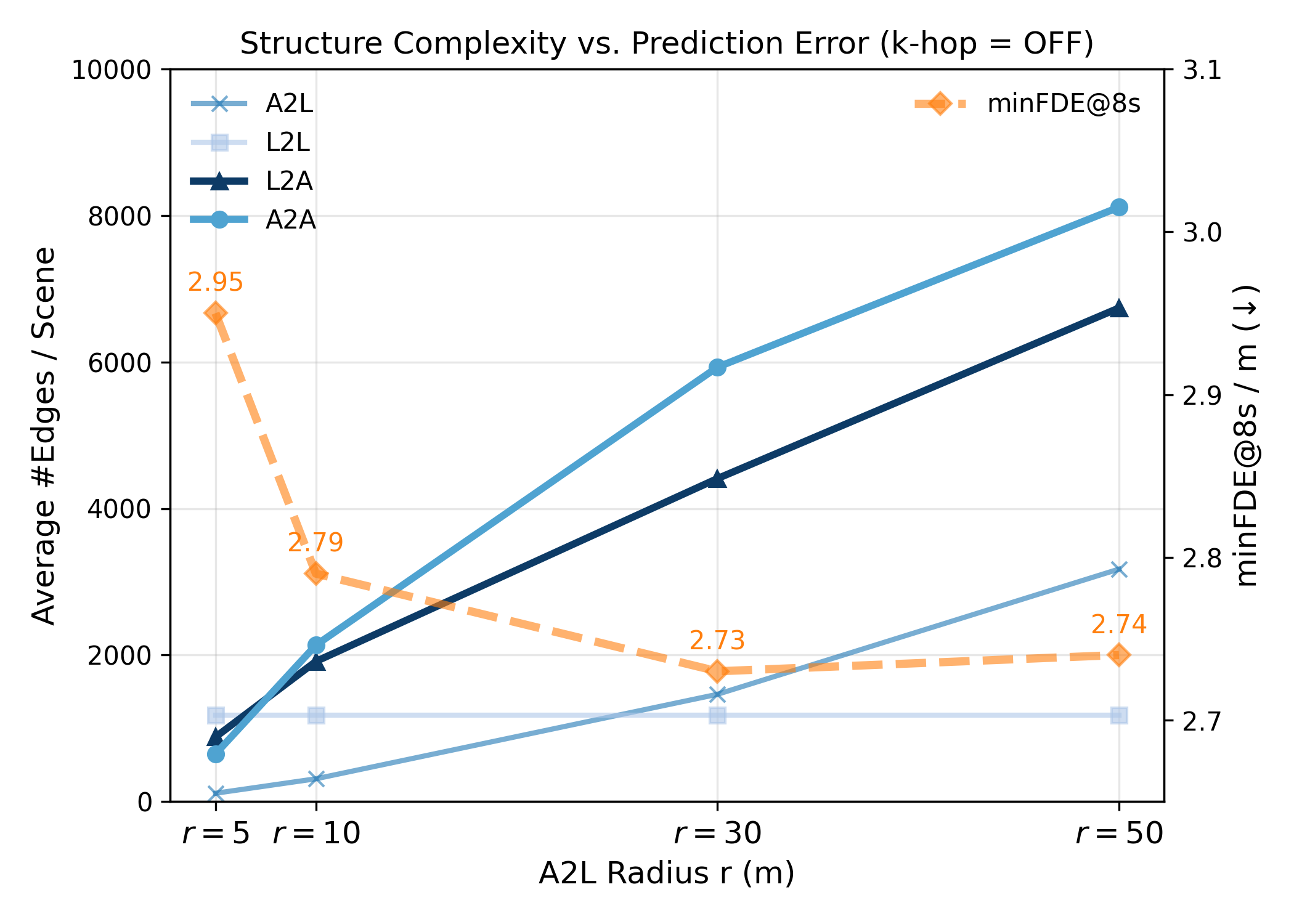}
    \caption{\textbf{Number of edges against A2L radius.} 
    Ablation of the A2L connection radius $r$ under a fixed OFF topology search strategy.
    The left y-axis shows the average number of edges per scene for different edge types (over 100 scenarios in the validation set), while the right y-axis reports the corresponding prediction error measured by minFDE@8s (over 41,000 scenarios in the validation set).
    Increasing $r$ leads to a rapid growth of higher-order interactions (L2A and A2A edges) induced by topological propagation, whereas the number of L2L edges remains unchanged.
    Prediction performance improves significantly for small $r$ but saturates beyond $r=10$, despite a continued rapid increase in graph complexity.
    }
    \label{fig: edges_vs_a2l_r_w_minFDE}
\end{figure}
\subsubsection{Effects of A2L Connection} 
\label{subsubsec: ablation_2}
\moxy{
The A2L connection radius $r$ determines the initial A2L edges in the scene graph and directly affects the scale of higher-order interactions induced by topological propagation (i.e., L2A and A2A edges). 
We conduct an ablation study with $r \in \{5, 10, 30, 50\}$ under a fixed OFF topology search strategy, and analyze both the resulting graph complexity and the prediction performance measured by minFDE@8s.
The results in Fig.~\ref{fig: edges_vs_a2l_r_w_minFDE} show that the number of L2L edges remains constant across different values of $r$, indicating that lane topology is independent of the geometric graph construction. 
Although the A2L edge count increases only mildly when $r$ changes from 5 to 10, the numbers of L2A and A2A edges exhibit a surge due to OFF propagation, revealing a clear nonlinear amplification effect of high-order interactions with respect to the initial geometric perturbation. 
A similar amplification pattern is \moa{also} observable when $r$ increases from 10 to 30, but with a reduced relative growth rate. When $r \geq 30$, the growth of A2A edges slows down.
\moa{From a performance perspective, minFDE@8s decreases significantly when increasing $r$ from 5 to 10, whereas further increasing $r$ from 10 to 30 and then to 50 yields only marginal improvements, despite the graph size continuing to increase rapidly.} 
This indicates that, once strong topological propagation is established, further enlarging the A2L radius mainly increases structural complexity rather than the discriminative power of the learned representations.
Based on the overall trade-off between graph complexity and prediction accuracy, we adopt $r=30$ as the A2L connection radius for benchmarking and $r=10$ for subsequent experiments on lane topology search strategy.
}

\begin{figure}[t]
    \centering
    \includegraphics[trim={0cm 0cm 0cm 0cm}, clip, width=0.48\textwidth]{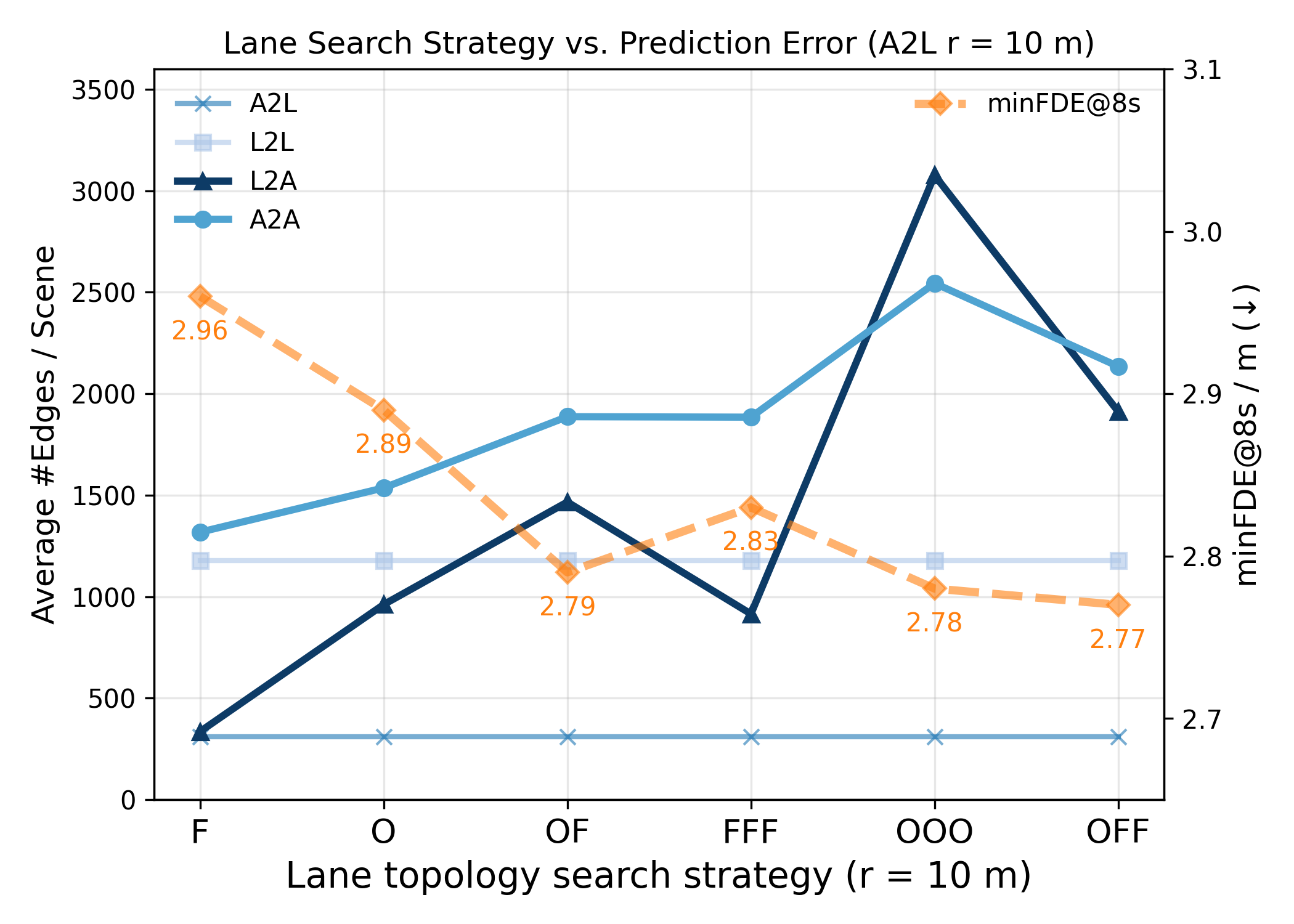}
    \caption{\textbf{Number of edges against lane topology search strategy.}
    Ablation of different lane topology search strategies under a fixed A2L connection radius $r = 10$.
    The left y-axis shows the average number of edges per scene for different edge types, while the right y-axis reports the corresponding prediction error measured by minFDE@8s.
    Purely forward strategies (FFF) yield sparse but incomplete interaction graphs, whereas fully omnidirectional search (OOO) introduces a large number of redundant higher-order interactions.
    In contrast, the proposed OFF strategy achieves the best trade-off by using significantly fewer edges than OOO while attaining competitive accuracy, leading to improved efficiency. Please note that this figure presents results from a separate ablation study.
    }
    \label{fig: edges_vs_topo_w_minFDE}
\end{figure}
\subsubsection{Effects of Lane Topology Search Strategy} 
\label{subsubsec: ablation_3}
\moxy{To highlight the structural and performance differences among lane topology search strategies, we fix the A2L radius to $r = 10 m$ in this ablation. 
As shown in Fig.~\ref{fig: edges_vs_topo_w_minFDE}, different combinations of omnidirectional (O) and forward (F) propagation lead to substantially different interaction structures and prediction accuracy. 
From the structural perspective, purely forward propagation (F) yields the sparsest L2A and A2A connections, while three-hop omnidirectional search (OOO) produces the densest interaction graph. 
The proposed OFF strategy achieves a significantly sparser structure than OOO while still maintaining sufficient interaction coverage. 
From the performance perspective, prediction accuracy does not monotonically improve with increasing graph density: 
\moa{Although OOO dramatically enlarges the L2A and A2A edge sets, its minFDE@8s is 2.78 meters, which is close to that of the much sparser OF and OFF. 
Notably, OFF achieves a minFDE@8s of 2.77, outperforming purely directional strategies (F and FFF) as well as the one-step omnidirectional strategy (O).} 
These results indicate that unconstrained dense topology expansion introduces redundant and noisy \moa{connections}, whereas forward-constrained expansion provides a stronger prior for realistic traffic modeling. 
By combining limited omnidirectional exploration with multi-hop forward expansions, OFF achieves a favorable trade-off between structural sparsity and interaction completeness, leading to higher computational efficiency while maintaining competitive prediction accuracy.}
\moa{OFF is adopted as the default strategy as it represents the minimal augmentation over OF that improves topological interaction coverage without introducing redundant complexity.}

\subsection{Visualizations}
\begin{figure*}[ht]
    \centering
    \begin{subfigure}{0.45\textwidth}
        \centering
        \includegraphics[trim={1cm 1cm 1cm 1cm}, width=\textwidth]{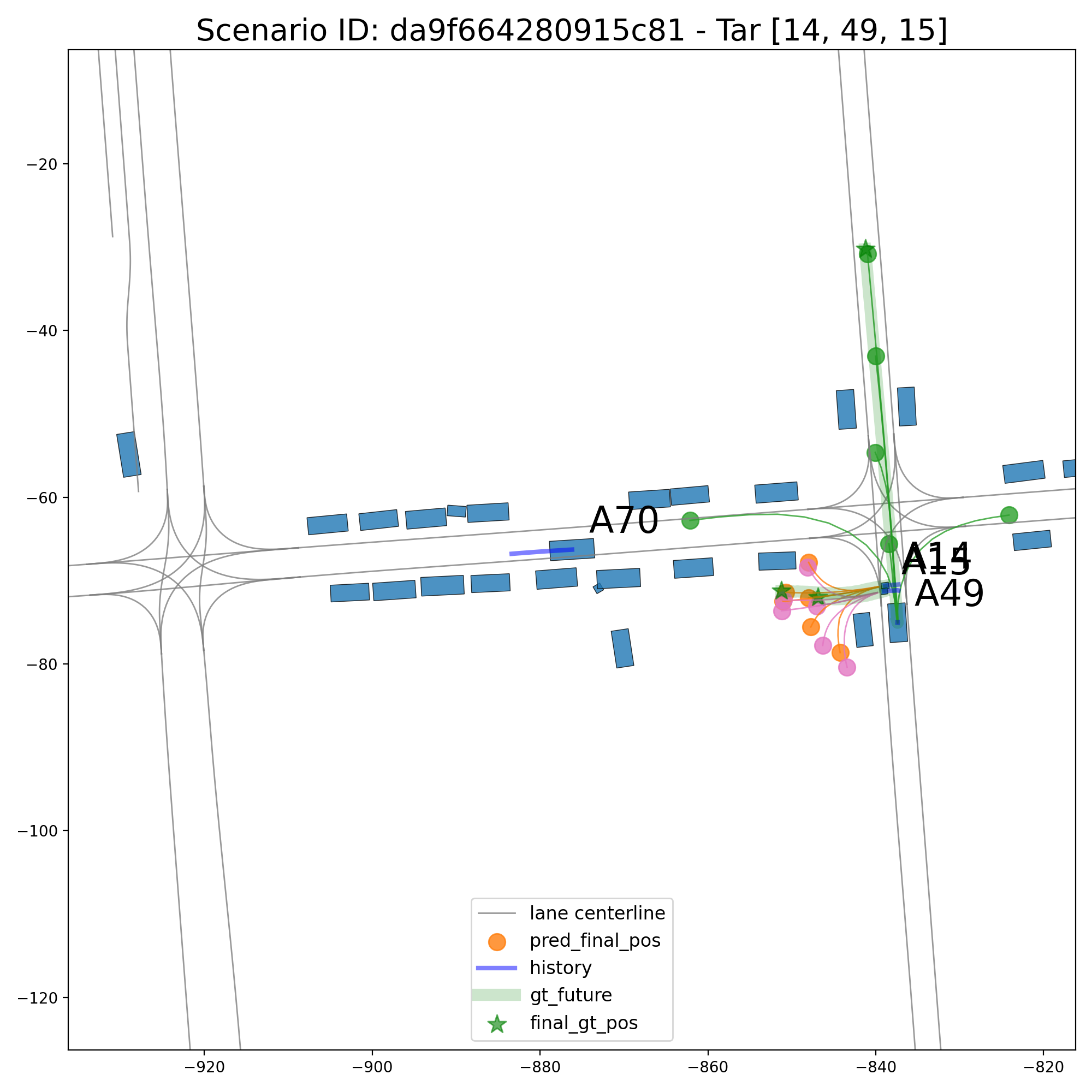}
        \caption{A14 and A15 are a pair of pedestrians crossing the road, while vehicle A49 has stopped to yield. The predicted trajectories of A14 and A15 exhibit similar motion patterns under comparable environmental conditions, indicating that the model effectively captures their shared behavioral characteristics. For vehicle A49, the prediction indicates that after yielding to the pedestrians, it will either proceed straight or execute a left or right turn, \wzf{aligning with} the underlying lane topology. }
        \label{fig:sub1}
    \end{subfigure}
    \hfill
    \begin{subfigure}{0.45\textwidth}
        \centering
        \includegraphics[trim={1cm 1cm 1cm 1cm}, width=\textwidth]{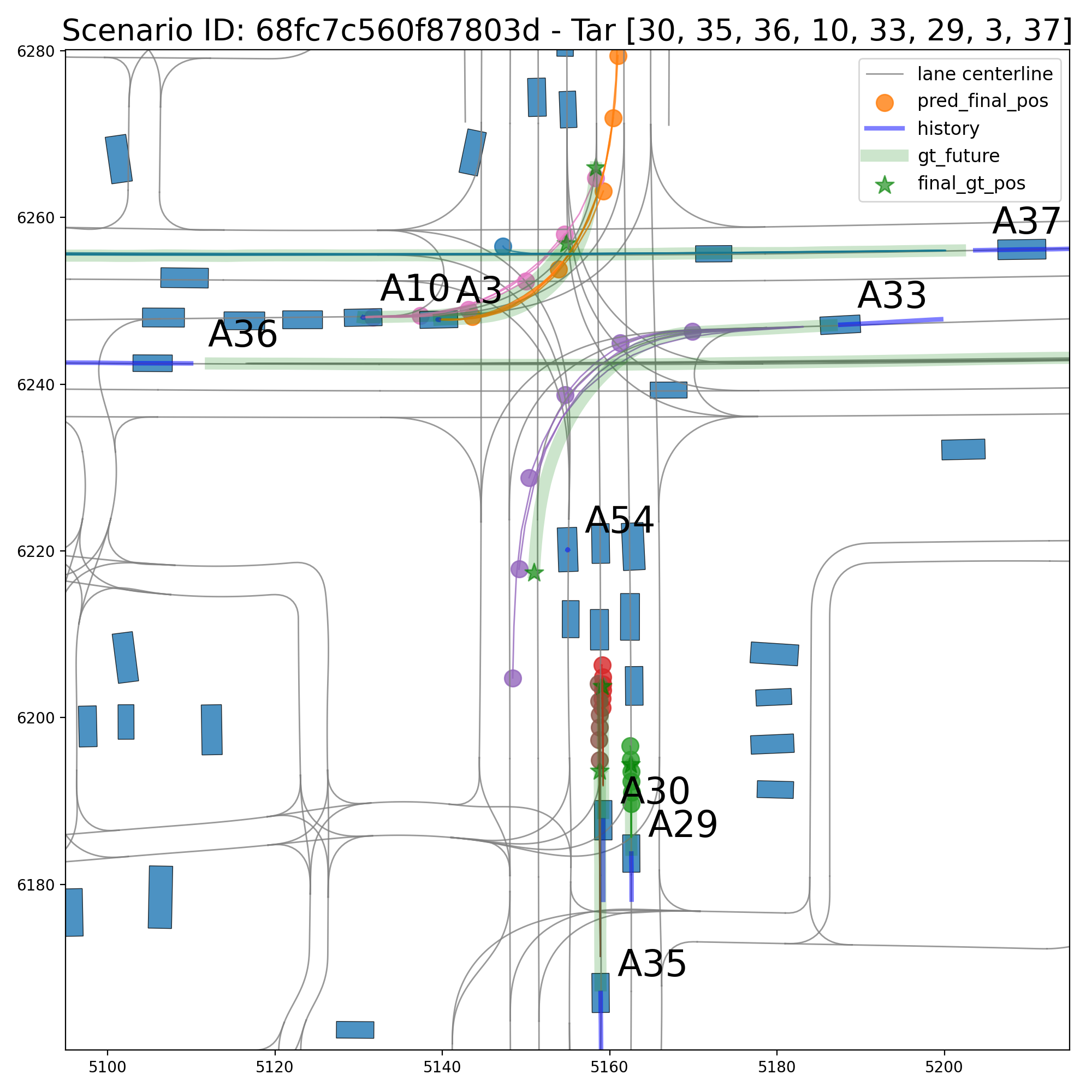}
        \caption{A29 and A30 slow down due to the preceding vehicles, highlighting the importance of interaction information. A36 and A37 traverse the intersection at relatively high speeds, whereas A3, A10, and A33 execute left turns in accordance with their local lane topologies. Interestingly, although the lane graph suggests that A33 has a U-turn option, SparScene does not predict this maneuver, as A33’s high speed makes it nearly infeasible.}
        \label{fig:sub2}
    \end{subfigure}
    \\[3mm]
    \begin{subfigure}{0.45\textwidth}
        \centering
        \includegraphics[trim={1cm 0cm 1cm 0cm}, width=\textwidth]{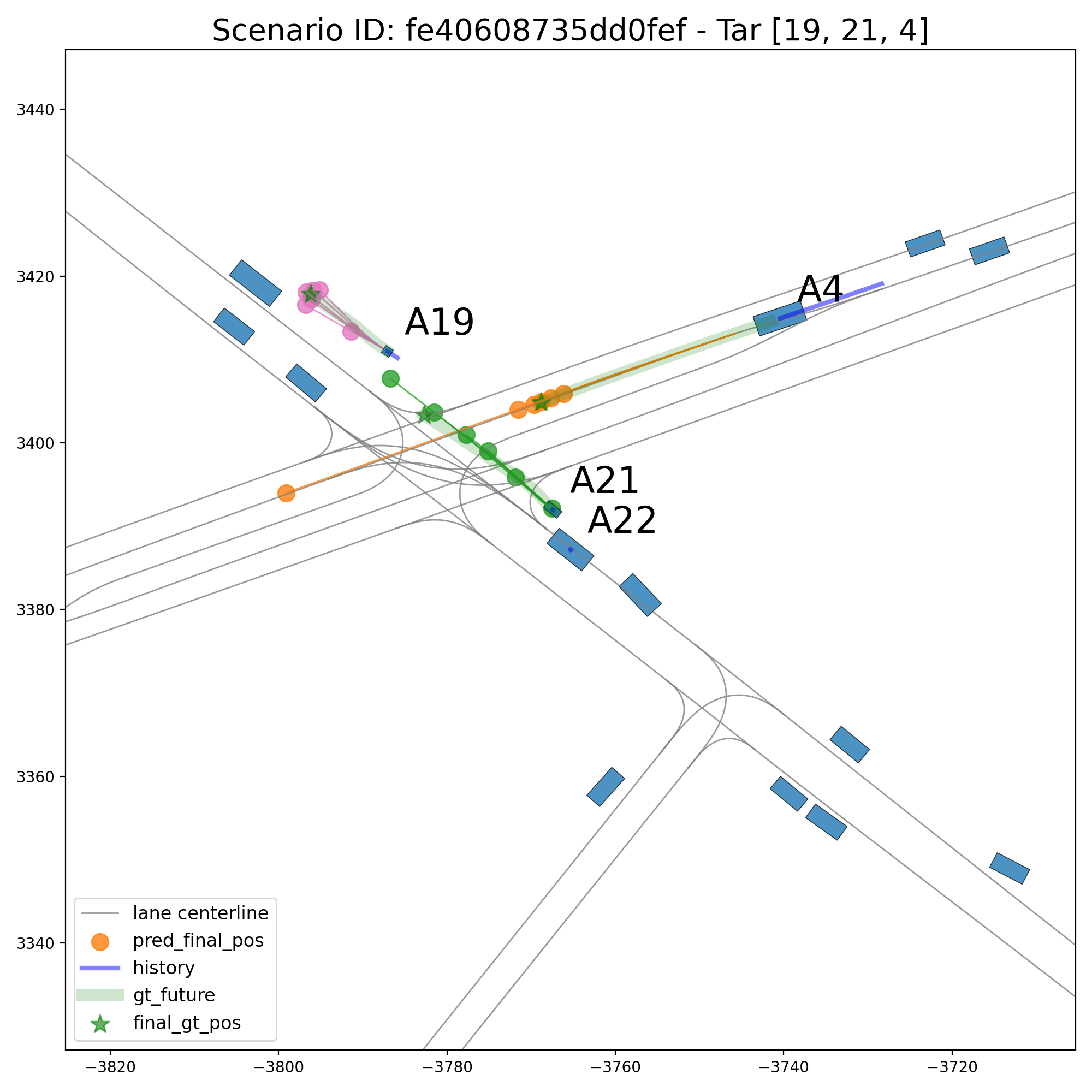}
        \caption{A19 is a pedestrian walking along the roadside, and A21 is a cyclist passing through the intersection. Vehicle A4 is approaching from the right, and its future motion may conflict with A21 at the intersection center. Our model anticipates this interaction risk and predicts that A4 is likely to yield to the cyclist, generating trajectories that slow down and ensure a safe passage. This demonstrates the model’s ability to infer multi-agent interactions and produce safety-aware motion predictions.}
        \label{fig:sub3}
    \end{subfigure}
    \hfill
    \begin{subfigure}{0.45\textwidth}
        \centering
        \includegraphics[trim={1cm 0cm 1cm 0cm}, width=\textwidth]{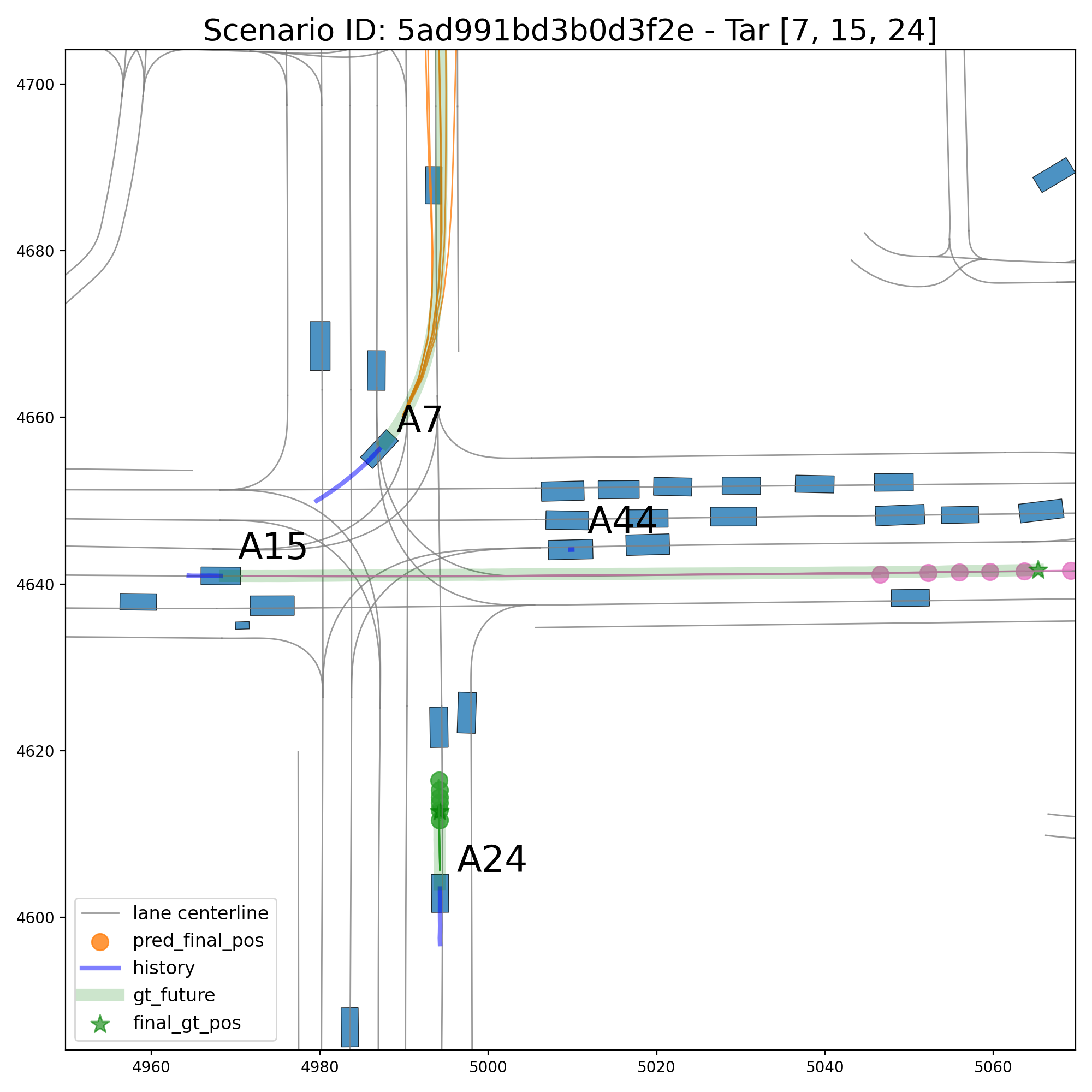}
        \caption{A15 is moving slowly at the moment, but it is expected to accelerate and proceed through the intersection, as the lane ahead is completely clear. Vehicle A7 performs a relatively fast left-turn maneuver and is \wzf{predicted} to continue along its intended path once the turn is completed. Conversely, A24 reduces its speed and eventually stops because a leading vehicle is stationary in its lane, prompting A24 to brake and maintain a safe distance to avoid a collision.}
        \label{fig:sub4}
    \end{subfigure}
    \vspace{3mm}
    \caption{\textbf{Visualization of MAMM predictions from SparScene.}}
    \label{fig:viz_preds}
\end{figure*}

\gjt{
To provide qualitative insights into the scene understanding capabilities of SparScene, we visualize four representative scenarios from the validation set as shown in Fig.~\ref{fig:viz_preds}. These cases are selected to span a broad spectrum of traffic complexities, covering diverse agent types, heterogeneous interaction patterns, and varying dependencies on dynamic states (\textit{dyn}), lane topology (\textit{lan}), and social interactions (\textit{int}).
Rather than cherry-picking samples with the lowest error metrics, we prioritize scenarios that highlight the model's reasoning process in challenging environments. The visualizations demonstrate that our model generates multimodal trajectories that are not only strictly compliant with map constraints but also responsive to multi-agent dynamics. Specific observations include:
\begin{itemize}[leftmargin=10pt]
    \item{\textbf{Heterogeneous Interaction and Temporal Reasoning. } As observed with Agent A49 in (a) and A4 in (c), the model successfully anticipates and responds to crossing pedestrians (A14/15 in (a)) and cyclists (A21 in (c)). Notably, the predicted strategies go beyond simple stopping; they exhibit complex temporal logic, such as a \textit{yield-then-resume} maneuver or anticipatory deceleration. This indicates that our representation effectively captures the causal dependencies between heterogeneous agents and encodes the temporal evolution of social interactions.}
    \item{
    \textbf{Topological Consistency and Motion Maintenance. } In routine navigation scenarios \wzf{without} traffic conflicts (e.g., lane keeping and turning for Agents A10, A3, A33 in (b) and A15, A7 in (d)), the predicted trajectories exhibit high kinematic consistency and strictly adhere to the lane centerlines. This confirms that our L2A module effectively injects map topology into the latent space, ensuring that the generated motions are valid and stable when map constraints dominate.
    }
    \item{\textbf{Multimodality and Edge-Case Reasoning. } SparScene generates diverse trajectories within reasonable topological and dynamic limits. A notable example is Agent A4 in (c): alongside the conservative yielding modes, the model identifies a plausible low-probability trajectory where the vehicle accelerates to pass before the pedestrian. This reflects the encoder's ability to model the full distribution of human behaviors, covering both compliant driving and aggressive, risk-aware possibilities.}
\end{itemize}
Across these scenarios, SparScene adapts the diversity of predicted futures to the underlying constraints: it produces interaction-driven branching modes under crossing conflicts, timing-dominated variations under queueing, and near-unimodal predictions when lane topology determines the motion. This adaptability indicates that the learned scene representation accurately captures the evolving semantic structure of traffic.
}

\section{Conclusion}
\label{sec: conclusion}
In this work, we presented SparScene, an efficient traffic scene representation framework designed for large-scale multi-agent trajectory generation. 
SparScene leverages HD-map lane topology to construct sparse and semantically meaningful connections between agents and lanes. 
By preserving full lane connectivity, modeling directed reachability on the lane graph, and adopting symmetric representations, SparScene enables compact, interpretable, and topology-compliant interaction modeling.
Built upon the proposed heterogeneous scene graph, we designed a lightweight topology-guided encoder that efficiently aggregates multimodal contextual information from surrounding agents and lanes without relying on densely connected or parameter-heavy architectures. 
SparScene naturally accommodates variable-sized traffic scenes and scales to thousands of traffic participants while maintaining millisecond-level inference latency and low memory cost. 
On the WOMD benchmark, SparScene achieves competitive prediction accuracy with significantly improved efficiency, demonstrating that high-performance trajectory generation does not require dense connections or expensive attention modeling, but rather an expressive and semantically grounded scene representation.

Overall, SparScene provides a principled solution for large-scale traffic scene representation and trajectory generation. 
We believe that the proposed structure-aware and symmetric representation offers meaningful insights into traffic interaction modeling and can serve as a foundation for future research on scalable trajectory prediction, multi-agent planning, behavior modeling, and large-scale traffic simulation. 
In future work, we plan to extend SparScene to multi-intersection coordination, incorporate signal phasing and right-of-way into the structural priors, and explore its integration into closed-loop simulators and planning frameworks.



\bibliographystyle{IEEEtran}
\bibliography{references.bib}

@inproceedings{liang2020learning,
  title={Learning lane graph representations for motion forecasting},
  author={Liang, Ming and Yang, Bin and Hu, Rui and Chen, Yun and Liao, Renjie and Feng, Song and Urtasun, Raquel},
  booktitle={European Conference on Computer Vision},
  pages={541--556},
  year={2020},
  organization={Springer}
}

@article{mo2025traffic,
  title={Traffic Scene Representation and Encoding With Graph Structure Learning and Exploration},
  author={Mo, Xiaoyu and Lou, Baichuan and Mao, Zhiqi and Huang, Qihang and Sun, Weigao and Wang, Yafei and Lv, Chen},
  journal={IEEE Transactions on Intelligent Transportation Systems},
  year={2025},
  publisher={IEEE}
}

@article{shi2024mtr++,
  title={Mtr++: Multi-agent motion prediction with symmetric scene modeling and guided intention querying},
  author={Shi, Shaoshuai and Jiang, Li and Dai, Dengxin and Schiele, Bernt},
  journal={IEEE Transactions on Pattern Analysis and Machine Intelligence},
  year={2024},
  publisher={IEEE}
}

@article{jia2023hdgt,
  title={Hdgt: Heterogeneous driving graph transformer for multi-agent trajectory prediction via scene encoding},
  author={Jia, Xiaosong and Wu, Penghao and Chen, Li and Liu, Yu and Li, Hongyang and Yan, Junchi},
  journal={IEEE transactions on pattern analysis and machine intelligence},
  year={2023},
  publisher={IEEE}
}

@article{konev2022motioncnn,
  title={Motioncnn: A strong baseline for motion prediction in autonomous driving},
  author={Konev, Stepan and Brodt, Kirill and Sanakoyeu, Artsiom},
  journal={arXiv preprint arXiv:2206.02163},
  year={2022}
}

@inproceedings{nayakanti2023wayformer,
  title={Wayformer: Motion forecasting via simple \& efficient attention networks},
  author={Nayakanti, Nigamaa and Al-Rfou, Rami and Zhou, Aurick and Goel, Kratarth and Refaat, Khaled S and Sapp, Benjamin},
  booktitle={2023 IEEE International Conference on Robotics and Automation (ICRA)},
  pages={2980--2987},
  year={2023},
  organization={IEEE}
}

@inproceedings{gu2021densetnt,
  title={Densetnt: End-to-end trajectory prediction from dense goal sets},
  author={Gu, Junru and Sun, Chen and Zhao, Hang},
  booktitle={Proceedings of the IEEE/CVF International Conference on Computer Vision},
  pages={15303--15312},
  year={2021}
}

@article{
  velickovic2018graph,
  title="{Graph Attention Networks}",
  author={Veli{\v{c}}kovi{\'{c}}, Petar and Cucurull, Guillem and Casanova, Arantxa and Romero, Adriana and Li{\`{o}}, Pietro and Bengio, Yoshua},
  journal={International Conference on Learning Representations},
  year={2018},
  url={https://openreview.net/forum?id=rJXMpikCZ},
}

@article{liu2025hybrid,
  title={Hybrid-prediction integrated planning for autonomous driving},
  author={Liu, Haochen and Huang, Zhiyu and Huang, Wenhui and Yang, Haohan and Mo, Xiaoyu and Lv, Chen},
  journal={IEEE Transactions on Pattern Analysis and Machine Intelligence},
  year={2025},
  publisher={IEEE}
}

@inproceedings{he2022masked,
  title={Masked autoencoders are scalable vision learners},
  author={He, Kaiming and Chen, Xinlei and Xie, Saining and Li, Yanghao and Doll{\'a}r, Piotr and Girshick, Ross},
  booktitle={Proceedings of the IEEE/CVF conference on computer vision and pattern recognition},
  pages={16000--16009},
  year={2022}
}

@article{sun2025speed,
  title={Speed always wins: A survey on efficient architectures for large language models},
  author={Sun, Weigao and Hu, Jiaxi and Zhou, Yucheng and Du, Jusen and Lan, Disen and Wang, Kexin and Zhu, Tong and Qu, Xiaoye and Zhang, Yu and Mo, Xiaoyu and others},
  journal={arXiv preprint arXiv:2508.09834},
  year={2025}
}

@article{chung2014empirical,
  title={Empirical evaluation of gated recurrent neural networks on sequence modeling},
  author={Chung, Junyoung and Gulcehre, Caglar and Cho, KyungHyun and Bengio, Yoshua},
  journal={arXiv preprint arXiv:1412.3555},
  year={2014}
}

@inproceedings{bergamini2021simnet,
  title={Simnet: Learning reactive self-driving simulations from real-world observations},
  author={Bergamini, Luca and Ye, Yawei and Scheel, Oliver and Chen, Long and Hu, Chih and Del Pero, Luca and Osi{\'n}ski, B{\l}a{\.z}ej and Grimmett, Hugo and Ondruska, Peter},
  booktitle={2021 IEEE International Conference on Robotics and Automation (ICRA)},
  pages={5119--5125},
  year={2021},
  organization={IEEE}
}

@inproceedings{suo2021trafficsim,
  title={Trafficsim: Learning to simulate realistic multi-agent behaviors},
  author={Suo, Simon and Regalado, Sebastian and Casas, Sergio and Urtasun, Raquel},
  booktitle={Proceedings of the IEEE/CVF Conference on Computer Vision and Pattern Recognition},
  pages={10400--10409},
  year={2021}
}

@inproceedings{ettinger2021large,
  title={Large scale interactive motion forecasting for autonomous driving: The waymo open motion dataset},
  author={Ettinger, Scott and Cheng, Shuyang and Caine, Benjamin and Liu, Chenxi and Zhao, Hang and Pradhan, Sabeek and Chai, Yuning and Sapp, Ben and Qi, Charles R and Zhou, Yin and others},
  booktitle={Proceedings of the IEEE/CVF International Conference on Computer Vision},
  pages={9710--9719},
  year={2021}
}

@article{mo2023map,
  title={Map-Adaptive Multimodal Trajectory Prediction Using Hierarchical Graph Neural Networks},
  author={Mo, Xiaoyu and Xing, Yang and Liu, Haochen and Lv, Chen},
  journal={IEEE Robotics and Automation Letters},
  year={2023},
  publisher={IEEE}
}

@INPROCEEDINGS {Argoverse,
  author = {Ming-Fang Chang and John W Lambert and Patsorn Sangkloy and Jagjeet Singh
       and Slawomir Bak and Andrew Hartnett and De Wang and Peter Carr
       and Simon Lucey and Deva Ramanan and James Hays},
  title = {Argoverse: 3D Tracking and Forecasting with Rich Maps},
  booktitle = {Conference on Computer Vision and Pattern Recognition (CVPR)},
  year = {2019}
}

@incollection{NEURIPS2019_9015,
title = {PyTorch: An Imperative Style, High-Performance Deep Learning Library},
author = {Paszke, Adam and Gross, Sam and Massa, Francisco and Lerer, Adam and Bradbury, James and Chanan, Gregory and Killeen, Trevor and Lin, Zeming and Gimelshein, Natalia and Antiga, Luca and Desmaison, Alban and Kopf, Andreas and Yang, Edward and DeVito, Zachary and Raison, Martin and Tejani, Alykhan and Chilamkurthy, Sasank and Steiner, Benoit and Fang, Lu and Bai, Junjie and Chintala, Soumith},
booktitle = {Advances in Neural Information Processing Systems 32},
editor = {H. Wallach and H. Larochelle and A. Beygelzimer and F. d\textquotesingle Alch\'{e}-Buc and E. Fox and R. Garnett},
pages = {8024--8035},
year = {2019},
publisher = {Curran Associates, Inc.},
url = {http://papers.neurips.cc/paper/9015-pytorch-an-imperative-style-high-performance-deep-learning-library.pdf}
}

@article{ngiam2021scene,
  title={Scene transformer: A unified multi-task model for behavior prediction and planning},
  author={Ngiam, Jiquan and Caine, Benjamin and Vasudevan, Vijay and Zhang, Zhengdong and Chiang, Hao-Tien Lewis and Ling, Jeffrey and Roelofs, Rebecca and Bewley, Alex and Liu, Chenxi and Venugopal, Ashish and others},
  journal={arXiv e-prints},
  pages={arXiv--2106},
  year={2021}
}

@article{shi2022motion,
  title={Motion transformer with global intention localization and local movement refinement},
  author={Shi, Shaoshuai and Jiang, Li and Dai, Dengxin and Schiele, Bernt},
  journal={Advances in Neural Information Processing Systems},
  volume={35},
  pages={6531--6543},
  year={2022}
}

@article{mo2022multi,
  title={Multi-agent trajectory prediction with heterogeneous edge-enhanced graph attention network},
  author={Mo, Xiaoyu and Huang, Zhiyu and Xing, Yang and Lv, Chen},
  journal={IEEE Transactions on Intelligent Transportation Systems},
  volume={23},
  number={7},
  pages={9554--9567},
  year={2022},
  publisher={IEEE}
}

@inproceedings{cui2019multimodal,
  title={Multimodal trajectory predictions for autonomous driving using deep convolutional networks},
  author={Cui, Henggang and Radosavljevic, Vladan and Chou, Fang-Chieh and Lin, Tsung-Han and Nguyen, Thi and Huang, Tzu-Kuo and Schneider, Jeff and Djuric, Nemanja},
  booktitle={2019 International Conference on Robotics and Automation (ICRA)},
  pages={2090--2096},
  year={2019},
  organization={IEEE}
}

@article{chai2019multipath,
  title={Multipath: Multiple probabilistic anchor trajectory hypotheses for behavior prediction},
  author={Chai, Yuning and Sapp, Benjamin and Bansal, Mayank and Anguelov, Dragomir},
  journal={arXiv preprint arXiv:1910.05449},
  year={2019}
}

@article{montali2023waymo,
  title={The waymo open sim agents challenge},
  author={Montali, Nico and Lambert, John and Mougin, Paul and Kuefler, Alex and Rhinehart, Nicholas and Li, Michelle and Gulino, Cole and Emrich, Tristan and Yang, Zoey and Whiteson, Shimon and others},
  journal={Advances in Neural Information Processing Systems},
  volume={36},
  pages={59151--59171},
  year={2023}
}

@article{zhao2024kigras,
  title={Kigras: Kinematic-driven generative model for realistic agent simulation},
  author={Zhao, Jianbo and Zhuang, Jiaheng and Zhou, Qibin and Ban, Taiyu and Xu, Ziyao and Zhou, Hangning and Wang, Junhe and Wang, Guoan and Li, Zhiheng and Li, Bin},
  journal={IEEE Robotics and Automation Letters},
  year={2024},
  publisher={IEEE}
}

@article{mo2023predictive,
  title={Predictive neural motion planner for autonomous driving using graph networks},
  author={Mo, Xiaoyu and Lv, Chen},
  journal={IEEE Transactions on Intelligent Vehicles},
  volume={8},
  number={2},
  pages={1983--1993},
  year={2023},
  publisher={IEEE}
}

@article{liu2025gatsim,
  title={GATSim: Urban Mobility Simulation with Generative Agents},
  author={Liu, Qi and Li, Can and Ma, Wanjing},
  journal={arXiv preprint arXiv:2506.23306},
  year={2025}
}

@inproceedings{sun2025sequence,
  title={Sequence Accumulation and Beyond: Infinite Context Length on Single GPU and Large Clusters},
  author={Sun, Weigao and Liu, Yongtuo and Tang, Xiaqiang and Mo, Xiaoyu},
  booktitle={Proceedings of the AAAI Conference on Artificial Intelligence},
  volume={39},
  number={19},
  pages={20725--20733},
  year={2025}
}

@article{wu2021flow,
  title={Flow: A modular learning framework for mixed autonomy traffic},
  author={Wu, Cathy and Kreidieh, Abdul Rahman and Parvate, Kanaad and Vinitsky, Eugene and Bayen, Alexandre M},
  journal={IEEE Transactions on Robotics},
  volume={38},
  number={2},
  pages={1270--1286},
  year={2021},
  publisher={IEEE}
}

@article{li2023survey,
  title={A survey on urban traffic control under mixed traffic environment with connected automated vehicles},
  author={Li, Jinjue and Yu, Chunhui and Shen, Zilin and Su, Zicheng and Ma, Wanjing},
  journal={Transportation research part C: emerging technologies},
  volume={154},
  pages={104258},
  year={2023},
  publisher={Elsevier}
}

@article{gao2024evaluation,
  title={Evaluation system for urban traffic intelligence based on travel experiences: A sentiment analysis approach},
  author={Gao, Sa and Ran, Qingsong and Su, Zicheng and Wang, Ling and Ma, Wanjing and Hao, Ruochen},
  journal={Transportation Research Part A: Policy and Practice},
  volume={187},
  pages={104170},
  year={2024},
  publisher={Elsevier}
}

@inproceedings{zeng2019end,
  title={End-to-end interpretable neural motion planner},
  author={Zeng, Wenyuan and Luo, Wenjie and Suo, Simon and Sadat, Abbas and Yang, Bin and Casas, Sergio and Urtasun, Raquel},
  booktitle={Proceedings of the IEEE/CVF conference on computer vision and pattern recognition},
  pages={8660--8669},
  year={2019}
}

@article{feng2023dense,
  title={Dense reinforcement learning for safety validation of autonomous vehicles},
  author={Feng, Shuo and Sun, Haowei and Yan, Xintao and Zhu, Haojie and Zou, Zhengxia and Shen, Shengyin and Liu, Henry X},
  journal={Nature},
  volume={615},
  number={7953},
  pages={620--627},
  year={2023},
  publisher={Nature Publishing Group UK London}
}

@inproceedings{huang2022recoat,
  title={Recoat: A deep learning-based framework for multi-modal motion prediction in autonomous driving application},
  author={Huang, Zhiyu and Mo, Xiaoyu and Lv, Chen},
  booktitle={2022 IEEE 25th International Conference on Intelligent Transportation Systems (ITSC)},
  pages={988--993},
  year={2022},
  organization={IEEE}
}

@inproceedings{gao2020vectornet,
  title={VectorNet: Encoding HD Maps and Agent Dynamics from Vectorized Representation},
  author={Gao, Jiyang and Sun, Chen and Zhao, Hang and Shen, Yi and Anguelov, Dragomir and Li, Congcong and Schmid, Cordelia},
  booktitle={Proceedings of the IEEE/CVF Conference on Computer Vision and Pattern Recognition},
  pages={11525--11533},
  year={2020}
}

@inproceedings{zhao2019multi,
  title={Multi-agent tensor fusion for contextual trajectory prediction},
  author={Zhao, Tianyang and Xu, Yifei and Monfort, Mathew and Choi, Wongun and Baker, Chris and Zhao, Yibiao and Wang, Yizhou and Wu, Ying Nian},
  booktitle={Proceedings of the IEEE Conference on Computer Vision and Pattern Recognition},
  pages={12126--12134},
  year={2019}
}

@inproceedings{alahi2016social,
  title={Social lstm: Human trajectory prediction in crowded spaces},
  author={Alahi, Alexandre and Goel, Kratarth and Ramanathan, Vignesh and Robicquet, Alexandre and Fei-Fei, Li and Savarese, Silvio},
  booktitle={Proceedings of the IEEE conference on computer vision and pattern recognition},
  pages={961--971},
  year={2016}
}

@article{zhang2026adversarial,
  title={Adversarial traffic scene generation considering harm, rarity, and ambiguity for autonomous driving testing},
  author={Zhang, Yiran and Lou, Shanhe and Lou, Baichuan and Zhang, Haitao and Lv, Chen},
  journal={Transportation Research Part C: Emerging Technologies},
  volume={182},
  pages={105426},
  year={2026},
  publisher={Elsevier}
}

@article{khandelwal2020if,
  title={What-if motion prediction for autonomous driving},
  author={Khandelwal, Siddhesh and Qi, William and Singh, Jagjeet and Hartnett, Andrew and Ramanan, Deva},
  journal={arXiv preprint arXiv:2008.10587},
  year={2020}
}

@inproceedings{zhou2022hivt,
  title={Hivt: Hierarchical vector transformer for multi-agent motion prediction},
  author={Zhou, Zikang and Ye, Luyao and Wang, Jianping and Wu, Kui and Lu, Kejie},
  booktitle={Proceedings of the IEEE/CVF conference on computer vision and pattern recognition},
  pages={8823--8833},
  year={2022}
}

@ARTICLE{mo2023map1,
  author={Mo, Xiaoyu and Liu, Haochen and Huang, Zhiyu and Li, Xiuxian and Lv, Chen},
  journal={IEEE Transactions on Intelligent Transportation Systems}, 
  title={Map-Adaptive Multimodal Trajectory Prediction via Intention-Aware Unimodal Trajectory Predictors}, 
  year={2023},
  volume={},
  number={},
  pages={1-13},
  doi={10.1109/TITS.2023.3331887}}

@inproceedings{deo2018convolutional,
  title={Convolutional social pooling for vehicle trajectory prediction},
  author={Deo, Nachiket and Trivedi, Mohan M},
  booktitle={Proceedings of the IEEE Conference on Computer Vision and Pattern Recognition Workshops},
  pages={1468--1476},
  year={2018}
 }

@inproceedings{Fey/Lenssen/2019,
  title={Fast Graph Representation Learning with {PyTorch Geometric}},
  author={Fey, Matthias and Lenssen, Jan E.},
  booktitle={ICLR Workshop on Representation Learning on Graphs and Manifolds},
  year={2019},
}

@inproceedings{lan2023sept,
  title={Sept: Towards efficient scene representation learning for motion prediction},
  author={Lan, Zhiqian and Jiang, Yuxuan and Mu, Yao and Chen, Chen and Li, Shengbo Eben},
  booktitle={The Twelfth International Conference on Learning Representations},
  year={2023}
}

@inproceedings{mo2020interaction,
  title={Interaction-aware trajectory prediction of connected vehicles using cnn-lstm networks},
  author={Mo, Xiaoyu and Xing, Yang and Lv, Chen},
  booktitle={IECON 2020 The 46th Annual Conference of the IEEE Industrial Electronics Society},
  pages={5057--5062},
  year={2020},
  organization={IEEE}
}

@inproceedings{salzmann2020trajectron++,
  title={Trajectron++: Dynamically-feasible trajectory forecasting with heterogeneous data},
  author={Salzmann, Tim and Ivanovic, Boris and Chakravarty, Punarjay and Pavone, Marco},
  booktitle={European Conference on Computer Vision},
  pages={683--700},
  year={2020},
  organization={Springer}
}

@inproceedings{vaswani2017attention,
  title={Attention is All you Need},
  author={Vaswani, Ashish and Shazeer, Noam and Parmar, Niki and Uszkoreit, Jakob and Jones, Llion and Gomez, Aidan N and Kaiser, Lukasz and Polosukhin, Illia},
  booktitle={NIPS},
  year={2017}
}

\end{document}